\PassOptionsToPackage{numbers,square,sort&compress}{natbib}
\PassOptionsToPackage{table,dvipsnames}{xcolor}
\documentclass[]{style/company_light}

\usepackage[utf8]{inputenc}
\usepackage{hyperref}
\usepackage{url}
\usepackage{booktabs}
\usepackage{amsfonts}
\usepackage{amsmath}
\usepackage{nicefrac}
\usepackage{microtype}
\usepackage{graphicx}
\usepackage{float}
\usepackage{subcaption}
\usepackage{multirow}
\usepackage{enumerate}
\usepackage{enumitem}
\usepackage[most]{tcolorbox}
\newtcolorbox{researchquestion}{enhanced, breakable=false, sharp corners, frame hidden,
    colback=brandprimary!15, borderline west={2.5pt}{0pt}{brandaccent},
    left=10pt, right=10pt, top=6pt, bottom=6pt, before skip=8pt, after skip=8pt,
    fontupper=\itshape, halign=center}
\usetikzlibrary{arrows.meta,positioning}
\usepackage{adjustbox}
\usepackage{array}
\usepackage{amssymb}
\usepackage{tabularx}
\usepackage{longtable}
\usepackage{listings}
\usepackage{needspace}

\makeatletter
\def\fps@figure{htbp}
\def\fps@table{htbp}
\makeatother

\newcommand{\compactfloatlayout}{%
  \renewcommand{\topfraction}{0.92}%
  \renewcommand{\bottomfraction}{0.88}%
  \renewcommand{\textfraction}{0.08}%
  \renewcommand{\floatpagefraction}{0.85}%
  \setcounter{topnumber}{3}%
  \setcounter{bottomnumber}{2}%
  \setcounter{totalnumber}{4}%
  \setlength{\textfloatsep}{10pt plus 2pt minus 2pt}%
  \setlength{\intextsep}{10pt plus 2pt minus 2pt}%
  \setlength{\floatsep}{8pt plus 2pt minus 2pt}%
  \interlinepenalty=0\relax
  \clubpenalty=10000\relax
  \widowpenalty=10000\relax
  \raggedbottom
}

\title{WorkflowArena: Benchmarking Executable Workflow Orchestration in LLM-Based Multi-Agent Systems}
\author[\dagger]{Zheng Jiang*}
\author{Fang Niu*}
\author{Bo Peng*}
\author{Houde Qian}
\author{Wenyi Li}
\author{Deyao Hong}
\author{Mingju Gao}
\author{Yizhe Chi}
\author{Youjie Zheng}
\author{Kaisen Yang}
\author[\ddagger]{Calvin Xiao}
\author[\ddagger]{Lifeng Sun}
\author[\ddagger]{Huan-ang Gao}

\affiliation{Navers Lab, Einsia.AI,\quad\quad Tsinghua University}
\contribution[*]{Equal Contribution}
\contribution[\dagger]{Project Lead}
\contribution[\ddagger]{Corresponding Authors}

\newcommand{\BenchmarkTasks}{145}
\newcommand{\BenchmarkDomains}{8}
\newcommand{\BenchmarkForms}{17}

\abstract{
Large language model agents are increasingly deployed as multi-agent systems rather than as a standalone model operating within a single context. Recent agent harnesses allow the model to instantiate worker agents and to author the executable workflow that coordinates them, shifting the design of system structure from developers to the model at run time. The quality of this organization therefore constitutes a distinct capability, yet existing benchmarks either measure end-to-end task completion or evaluate coordination under structures specified in advance. We formalize this capability as \emph{workflow orchestration} and introduce \emph{WorkflowArena}, a benchmark of \BenchmarkTasks{} long-horizon tasks spanning \BenchmarkDomains{} professional domains, from software engineering and formal mathematics to legal analysis and 3D production. Each task specifies an initial workspace and an integrated deliverable composed of interdependent components, which is assessed by a hidden verifier. Agents determine autonomously whether and how to delegate, and all delegation is recorded, enabling joint evaluation of deliverable quality and the organization of the underlying work. Experiments with nine models show that no single configuration performs best across all domains and that systems exhibit distinct delegation strategies: at high effort, Claude models in Claude Code invoke workflows less frequently but with larger worker groups, whereas GPT models in Codex delegate more frequently in smaller units. When the executing model is held fixed on a challenging subset of tasks, workflows authored by Claude Opus 5.5 substantially improve weaker executors, indicating that workflow organization is itself a measurable determinant of task performance.
}

\date{\today}
\metadata[Homepage]{\url{https://lab.einsia.ai/workflowarena}}
\correspondence{\email{nana@einsia.ai}}

\begin{document}

\maketitle

\compactfloatlayout
\section{Introduction}
Large language models have evolved from text generators into agents that act in external environments and pursue goals over long horizons \citep{yao2022react,shinn2023reflexion,yang2024sweagent}. As tasks grow longer, the limiting factor shifts from what a single agent can accomplish to how the work is organized. Long-horizon tasks consist of interdependent subtasks that call for different expertise and independent verification, and a single context that must retain the entire history of such work soon becomes a bottleneck. Further progress therefore depends on \emph{system intelligence}, the ability of an agent system to organize multiple intelligent components into a coherent and adaptive whole \citep{feng2026graphengineering}. Current systems already delegate and replan dynamically \citep{fourney2024magenticone}, and recent harnesses allow the model to author the executable workflow that coordinates its workers \citep{anthropic2026dynamicworkflows,liu2026dynaflow}. The structure of a multi-agent system, once fixed by its developers, is thus increasingly determined by the model at run time, and its quality becomes a property of the model that requires evaluation in its own right.

We refer to this capability as \emph{workflow orchestration}. A workflow is an executable specification of how a task is partitioned among agents, which context each agent receives, and how intermediate results are verified and integrated. These decisions shape system behavior beyond the control of any individual agent. Running independent units in parallel, for instance, shortens the critical path but adds coordination cost, whereas an explicit verification point can catch an error before it propagates to downstream work \citep{anthropic2024effectiveagents,kim2023llmcompiler,anthropic2026dynamicworkflows}. The appropriate organization depends on the dependency structure of the task and on the available budget. Because the same agents can succeed or fail depending on how their work is organized, a premise we test directly in \S\ref{sec:workflow-transfer}, orchestration is a system-level capability rather than an implementation detail. This leaves a central question that current evaluations do not answer:

\begin{researchquestion}
To what extent can LLM agents autonomously orchestrate multi-agent workflows\\ to accomplish realistic long-horizon tasks?
\end{researchquestion}

\begin{figure}[t]
    \centering
    \includegraphics[width=\linewidth]{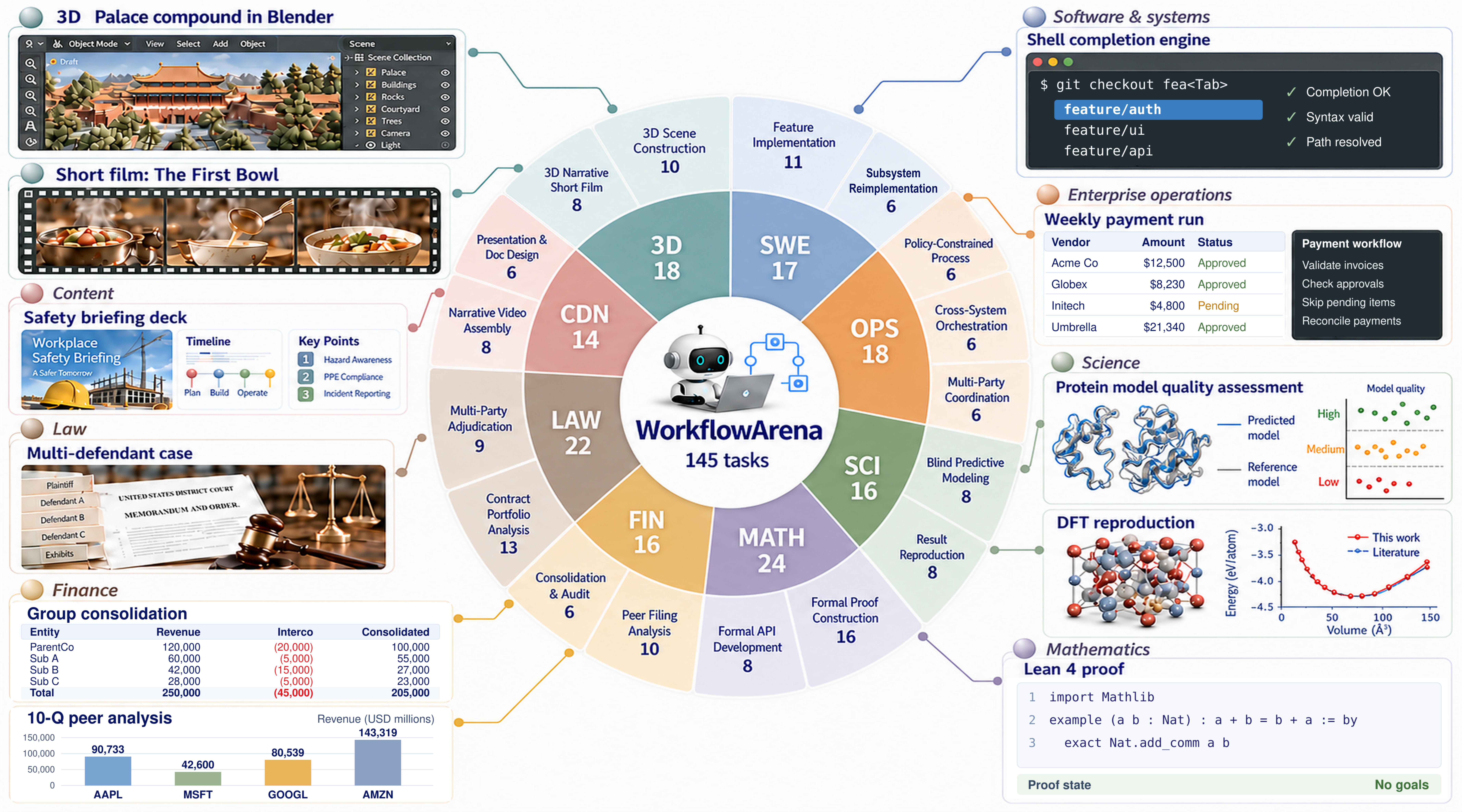}
    \caption{WorkflowArena comprises \BenchmarkTasks{} tasks across \BenchmarkDomains{} domains and \BenchmarkForms{} task forms. The inner ring shows domain totals, the outer ring shows task-form counts, and the surrounding panels illustrate representative deliverables.}
    \label{fig:coverage}
\end{figure}

Existing benchmarks address parts of this question. End-to-end agent benchmarks measure task completion in realistic environments, including long-horizon and compositional work \citep{liu2023agentbench,zhou2024webarena,xie2024osworld,trivedi2024appworld,jimenez2024swebench,boisvert2024workarenapp}, but they treat the agent as a single unit, so how its work is divided remains invisible. Multi-agent benchmarks study collaboration under predefined coordination protocols or among independently tasked agents \citep{zhu2025multiagentbench,mediratta2026parallact}; the structure under evaluation is chosen by the benchmark designer rather than by the model. Workflow benchmarks evaluate generated plans and workflow structures \citep{xiao2024flowbench,qiao2025worfbench,ren2026orchbench}, typically offline or in simulation, so the decisions they score never play out on real artifacts. None of them observes a workflow that the model writes itself and then adapts as its workers report back on real work.

We introduce \emph{WorkflowArena} to evaluate this capability directly. Evaluating orchestration requires tasks in which the way work is organized changes the result, and a protocol that leaves this organization to the model while recording it and charging its cost against the task. WorkflowArena provides both. Its \BenchmarkTasks{} tasks span \BenchmarkDomains{} work domains and \BenchmarkForms{} task forms (Figure~\ref{fig:coverage}) and are built from versioned professional artifacts or from authored scenarios whose dependencies are fully specified. Each task requests a deliverable composed of many separable units that depend on one another, such as clause-level analyses that must be reconciled across a contract portfolio or proofs that build on earlier results, so decisions about decomposition and integration change the outcome. A hidden verifier scores the submission against the task contract, including checks that individually correct but unreconciled components cannot pass. The agent decides whether to delegate at all, and the time spent organizing the work is drawn from the same budget as the work itself. Because every delegation passes through a recorded interface, each outcome can be read together with the workflow that produced it.

Our empirical analysis operates at two levels. At the system level, we evaluate 25 configurations of nine models running in Claude Code or Codex. The strongest configuration, Claude Opus 5.5 with Claude Code at max effort, resolves only 36.95\% of tasks on average across domains, and no configuration performs best in every domain. Execution traces reveal distinct delegation strategies: at high effort, Claude Opus 5.5 and Claude Sonnet 5.5 with Claude Code launch fewer workflows than GPT-6 Astra and GPT-6.1 Sol with Codex, but each workflow contains more workers. Because system-level results combine orchestration with the capability of the agents that carry out the work, we also hold the executor fixed: a source model writes and revises the workflow without editing the workspace, and every worker is an instance of a weaker execution model. On a challenging subset of tasks, workflows authored by Opus raise the continuous scores of all four execution models, by 6.41 points on average in the static mode, whereas workflows from Astra, itself far stronger than these executors, help them far less and lower the scores of two of them. Orchestration quality is thus a measurable source of performance that appears distinct from the ability to complete the task alone.

Our contributions are:
\begin{itemize}[leftmargin=*]
    \item \textit{Benchmark.} We construct WorkflowArena, \BenchmarkTasks{} long-horizon tasks across \BenchmarkDomains{} domains and \BenchmarkForms{} task forms, in which each deliverable contract is paired with domain-specific hidden verification.
    \item \textit{Evaluation protocol.} We route every delegation through a recorded interface under a shared time budget while leaving the decision to delegate to the agent, so that its organization of work can be analyzed alongside deliverable outcomes and resource use.
    \item \textit{Empirical findings.} We compare 25 model--effort configurations by resolved rate and delegation behavior, and use a fixed-executor transfer experiment to isolate the contribution of the workflow and distinguish orchestration quality from solo task performance.
\end{itemize}

\section{Related Work}

\paragraph{Long-horizon agent benchmarks.}
Agent evaluation has shifted toward interactive environments with outcome-based verification. AgentBench, WebArena, OSWorld, AppWorld, and SWE-bench ground evaluation in heterogeneous environments, websites, operating systems, applications, and software repositories \citep{liu2023agentbench,zhou2024webarena,xie2024osworld,trivedi2024appworld,jimenez2024swebench}. GAIA and AssistantBench assess tool-using assistants on realistic web tasks, while WorkArena and WorkArena++ study enterprise software use and compositional workflows \citep{mialon2023gaia,yoran2024assistantbench,drouin2024workarena,boisvert2024workarenapp}. TheAgentCompany, MLE-bench, and ScienceAgentBench extend evaluation to workplace tasks, machine learning engineering, and scientific programming \citep{xu2024theagentcompany,chan2024mlebench,chen2024scienceagentbench}, and the $\tau$-bench family evaluates policy-constrained interaction and coordination in shared environments \citep{yao2024taubench,barres2025tau2bench}. Many of these tasks already involve long horizons and interdependent steps, but the agent is scored as a single unit. WorkflowArena retains their executable environments and independent verification while making the organization of the work observable, since every delegation the agent makes is recorded alongside its deliverable.

\paragraph{Multi-agent system design and automation.}
MetaGPT, ChatDev, and AutoGen organize collaboration through specialized roles and programmable agent conversations \citep{hong2023metagpt,qian2023chatdev,wu2023autogen}. MultiAgentBench compares coordination protocols, ParallACT studies conflicts among independently tasked agents in shared codebases, and MAST categorizes failure modes of existing systems \citep{zhu2025multiagentbench,mediratta2026parallact,cemri2025mast}. A second line of work automates system design by searching over computational graphs, code-defined agents, executable workflows, and modular architectures \citep{zhuge2024gptswarm,hu2024adas,zhang2024aflow,shang2024agentsquare}; MaAS samples query-dependent systems from a learned supernet \citep{zhang2025maas}, and MAS-Loop refines generated systems using execution trajectories \citep{lin2026masloop}. ADAS and AFlow further report that the designs they discover transfer across models \citep{hu2024adas,zhang2024aflow}. Other systems adapt their structure at run time: DyLAN dynamically selects and connects agents, Magentic-One delegates and replans during execution, and DynaFlow generates executable workflow programs \citep{liu2023dylan,fourney2024magenticone,liu2026dynaflow}. These methods produce structures through design procedures, fixed controllers, or trained orchestrators, and assess them through downstream task accuracy. WorkflowArena instead treats the structure as an output of the model under evaluation, and its transfer experiment holds the executor fixed so that the contribution of a workflow can be measured separately from the model that authored it.

\paragraph{Workflow generation and evaluation.}
ReAct and Reflexion interleave execution with observation and feedback, ReWOO separates planning from tool observations, and LLMCompiler schedules function calls in parallel \citep{yao2022react,shinn2023reflexion,xu2023rewoo,kim2023llmcompiler}. FlowBench studies planning with supplied workflow knowledge, WorFBench evaluates generated sequence and graph structures together with their downstream utility, and OrchBench isolates orchestration-plan quality through deterministic simulation and validates its scores against real executions \citep{xiao2024flowbench,qiao2025worfbench,ren2026orchbench}. SkillsBench evaluates procedural knowledge by comparing matched conditions with and without skills \citep{li2026skillsbench}. These studies separate the quality of a workflow structure from its realized utility, but they evaluate structures largely offline or in simulation. Interface design and simple execution pipelines can also materially affect task performance, as SWE-agent and Agentless illustrate \citep{yang2024sweagent,xia2024agentless}, which motivates evaluating the model together with its harness. WorkflowArena evaluates workflows as they execute on professional artifacts, where the model revises its plan from the results its workers return and the time spent on orchestration counts against the task.

\section{The WorkflowArena Benchmark}
\label{sec:benchmark}

WorkflowArena evaluates agents that organize their own work on long-horizon tasks. A benchmark for this capability must satisfy five requirements. First, organization must be \emph{consequential}: tasks must contain enough separable and interdependent work that decisions about decomposition and integration change the outcome. Second, it must be \emph{autonomous}, since a benchmark that prescribed roles or a workflow would measure compliance with that structure rather than the ability to choose one. Third, evaluation must be \emph{integration-sensitive}, so that individually correct but uncoordinated components do not earn full credit. Fourth, the cost of organizing must be \emph{accounted for}, so that every organizational step competes with the work itself for time. Fifth, the organization must be \emph{observable}, so that each outcome can be related to the workflow that produced it. After formalizing the task (\S\ref{sec:formulation}), we describe how the task design satisfies the first and third requirements (\S\ref{sec:construction}) and how the evaluation protocol satisfies the remaining three (\S\ref{sec:protocol}). The transfer experiment in \S\ref{sec:workflow-transfer} complements these properties by separating the contribution of a workflow from the capability of the agents that execute it.

\subsection{Task formulation}
\label{sec:formulation}

A WorkflowArena task is a tuple
\begin{equation}
    \tau = (s_0, g, \mathcal{U}, \mathcal{E}, \mathcal{B}, V),
\end{equation}
where $s_0$ is the initial environment state, $g$ is the goal stated to the agent, $\mathcal{U}$ is the public action and tool surface, $\mathcal{E}$ is the set of task artifacts and execution environments, $\mathcal{B}$ is the resource budget, and $V$ is an independent verifier. The agent observes the instruction and the public part of $(s_0,\mathcal{U},\mathcal{E},\mathcal{B})$, but never the verifier, gold values, or hidden tests.

During execution, the agent makes the organizational decisions that define a workflow, and these decisions are realized through tool calls and workspace updates recorded in the execution history $h_T$. Execution within budget $\mathcal{B}$ yields a terminal state $s_T$ containing the final artifact, which the verifier evaluates together with any task-relevant state transitions:
\begin{equation}
    r = V(s_T,h_T) \in [0,1].
\end{equation}
The reward depends on the task outcome and on any required process constraints, but not on the number of workers or the shape of the workflow. A single sequential agent is therefore a valid workflow. The benchmark rewards whichever organization suits the task rather than presupposing that additional agents or parallelism are beneficial, which leaves the choice of organization to the model.

\begin{figure}[t]
    \centering
    \includegraphics[width=\linewidth]{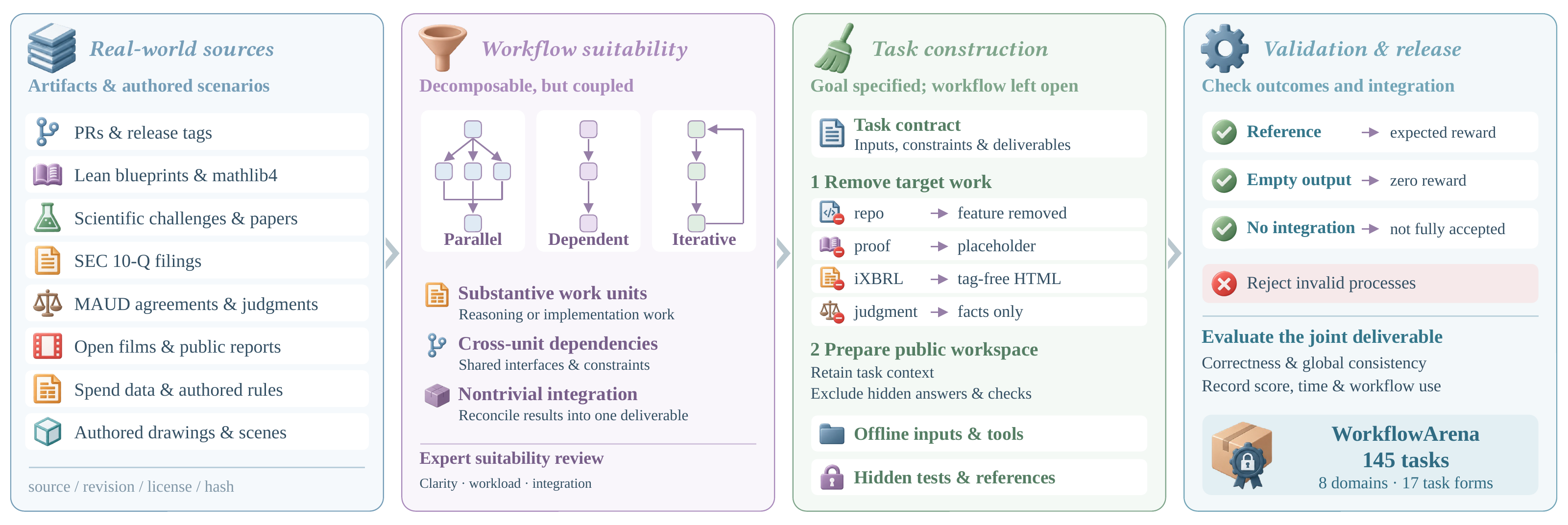}
    \caption{WorkflowArena benchmark construction. From left to right: real-world materials and authored scenarios provide task sources; suitability screening selects substantial, coupled work requiring integration; task construction specifies goals and constraints while leaving the workflow open, and separates public inputs from hidden references; validation checks reference solutions and failing submissions, with evaluation focused on the joint deliverable. The parallel, dependent, and iterative diagrams illustrate possible task structures rather than prescribed agent workflows.}
    \label{fig:construction}
\end{figure}

\Needspace{8\baselineskip}
\subsection{Task design and construction}
\label{sec:construction}

WorkflowArena contains \BenchmarkTasks{} tasks organized at two levels (Figure~\ref{fig:coverage}; Table~\ref{tab:suite} in Appendix~\ref{app:task_catalog}). The \BenchmarkDomains{} domains identify the work setting, ranging from software and enterprise operations to law and 3D production, and the \BenchmarkForms{} task forms identify the deliverable and its verification structure within each domain. Tasks are selected for three properties that together make organization consequential. A substantial workload supplies separable units of work that can be assigned to different workers. Dependencies or shared constraints among these units make their ordering and integration matter. Scoring criteria defined on the integrated deliverable ensure that these decisions are reflected in the result. The properties constrain what the agent must deliver while leaving decomposition and execution order to the agent, as summarized in Figure~\ref{fig:construction}.

\paragraph{Task materials.}
Tasks are built from real professional materials, from open-source repositories and formal mathematics libraries to corporate filings, legal agreements, and media assets; Table~\ref{tab:suite} lists the sources for each task form. These materials supply the scale, structure, and domain conventions of the work. Where a setting requires controlled state transitions or exceptions, as in operational processes and multi-entity accounting, the materials are embedded in authored scenarios whose dependencies are fully specified.

\paragraph{Interdependent work.}
Interdependence gives decomposition and integration their consequences, and it is expressed as requirements on the final deliverable. In a contract-portfolio task, each agreement can be analyzed clause by clause, yet the deliverable also requires relations and conflicts across contracts, so separately produced analyses must be reconciled into one consistent report. In a multi-entity consolidation task, adjustments to each entity propagate through intercompany balances into the consolidated ledger. In formal proof construction, the target proofs belong to a development in which later results rely on earlier ones, and in 3D scene construction, separately built components must meet at shared spatial interfaces. Predictive modeling instead rewards exploring alternative approaches that are later compared and combined. In every case, units may be developed separately, but the evaluation examines how they fit together in the submitted result.

\paragraph{Instantiation and acceptance.}
Each task specifies an initial workspace, a target deliverable, and task-specific constraints. For reconstruction tasks, the target work is removed while the surrounding context is retained. The task contract states the requirements on the deliverable, whereas reference outputs and hidden checks are excluded from the agent-facing workspace. Acceptance is defined on the final artifact. Wherever units of work depend on one another, the contract includes checks that only an integrated result can pass, such as regression tests in software or reconciliation in accounting. Operational tasks are additionally evaluated on their action histories, because a correct final state can be reached through an invalid process.

\paragraph{Review and verifier testing.}
Human experts review the instructions, materials, and scoring criteria of each task, and tasks with ambiguous requirements, defective materials, or no substantive integration are excluded. Each retained verifier is tested against reference solutions and against submissions that should fail. Among the latter, submissions with correct components but no integration keep the evaluation integration-sensitive, since delegated components that are never reconciled cannot pass; submissions that reach a correct outcome through an invalid process play the same role for process constraints. Appendix~\ref{app:task_catalog} details the source material and instantiation of each task form.

\subsection{Evaluation protocol}
\label{sec:protocol}

We evaluate complete model--harness configurations under a common end-to-end wall-clock budget of four hours. Each task starts from a clean container workspace, and the budget is shared by the main agent and its workers. It covers all activity after the task starts, whereas environment preparation and grading lie outside it. Because the budget is shared, organization has a cost: parallel workers can shorten the critical path, but every step needed to coordinate them consumes the same time as the work itself. Task outcomes therefore reflect deliverable quality under a fixed time constraint, while token consumption and estimated inference cost characterize the resources each configuration uses.

\paragraph{Agent interface.}
The agent has full access to the task workspace, including the shell, tests, and direct file editing. It is also given a Workflow tool that launches background workflows with up to 16 concurrent workers and allows the agent to inspect, stop, and revise them. We write \emph{Workflow} for this tool and its invocations, and \emph{workflow} for the general concept. Each worker is a fresh instance of the agent itself, with the same model, harness, and effort. Because workers start from a fresh context and share only the workspace, the agent must decide what each worker needs to know, which makes context allocation part of the evaluated behavior. Native sub-agent mechanisms are disabled, so every delegation passes through the Workflow tool and is recorded. Using the tool remains optional, which leaves the decision to delegate entirely to the agent. The agent may revise its workflow in response to public execution feedback until it stops or exhausts the budget, after which the verifier scores the final artifact. Each execution thus yields a trace of Workflow invocations, worker executions, and revisions, together with the time and tokens consumed by the main agent and each worker.

\paragraph{Scoring and aggregation.}
Each task receives a continuous score $S_t=100r_t\in[0,100]$, where $r_t$ is the verifier reward defined in \S\ref{sec:formulation}. The score reflects how well the submitted artifact and execution satisfy the task requirements, including functional correctness, integration, and process constraints where applicable; Appendix~\ref{app:scoring} describes the task-specific criteria. Every configuration is evaluated on all \BenchmarkTasks{} tasks. With $T_d$ denoting the set of tasks in domain $d$, the domain score is
\begin{equation}
    \bar S_d=\frac{1}{|T_d|}\sum_{t\in T_d}S_t,
\end{equation}
and the Overall Score is the equal-weight mean of the \BenchmarkDomains{} domain scores. A task is \emph{resolved} when its submission satisfies all acceptance criteria of its domain (Appendix~\ref{app:pass-rates}). With $n^{\mathrm{res}}_d$ resolved tasks in domain $d$, the domain resolved rate is $\mathrm{RR}_d=100\,n^{\mathrm{res}}_d/|T_d|$, and the Overall Resolved, the equal-weight mean of the \BenchmarkDomains{} domain resolved rates, serves as our primary measure. The Overall Score complements it by crediting partial progress and is reported in Appendix~\ref{app:score-resources}. Agent errors, missing deliverables, and budget exhaustion are not treated separately; the resulting submission is scored under the same task-specific rules.

\section{Experiments}
\label{sec:experiments}

The experiments address the research question from three perspectives. After describing the setup (\S\ref{sec:setup}), we measure how far agents accomplish the tasks through task performance (\S\ref{sec:main-results}) and resource use (\S\ref{sec:resources}). We then examine how they organize the work, characterizing delegation across systems, effort settings, and domains (\S\ref{sec:workflow-usage}). Because system-level outcomes combine orchestration with execution capability, a transfer experiment holds the executor fixed to isolate the contribution of the workflow itself (\S\ref{sec:workflow-transfer}). Finally, three 3D construction tasks illustrate in detail what the resulting workflows look like (\S\ref{sec:case-study}).

\subsection{Experimental setup}
\label{sec:setup}

\paragraph{Evaluated systems.}
Table~\ref{tab:main} covers nine models and 25 model--effort configurations. Claude Opus 5.5 and Claude Sonnet 5.5 run in Claude Code, and GPT-6 Astra and GPT-6.1 Sol run in Codex; these four are evaluated at low, medium, high, xhigh, and max effort. GPT-6 Luna runs in Codex at max effort. Kimi K3.1 Pro, GLM-5.3 Flash, and DeepSeek V4.1 Flash run in Claude Code at max effort, and Doubao-Seed-Evolving runs in Claude Code at high effort. Each configuration denotes the complete model--harness system, with effort set through the harness. We hereafter refer to the models as Opus, Sonnet, Astra, Sol, Luna, Kimi, GLM, DeepSeek, and Doubao.

\paragraph{Scoring and analysis.}
All configurations are evaluated on the same \BenchmarkTasks{} tasks under the protocol in \S\ref{sec:protocol}. Content and 3D artifacts are graded against task rubrics by Gemini-3.8-Flash, while the remaining domains use programmatic verifiers (Appendix~\ref{app:scoring}). Resource summaries include both the main agent and its workers; token totals count input and output, including cached input, normalized across providers as described in Appendix~\ref{app:execution}, and inference cost is estimated from recorded token usage and model-specific API prices. The delegation analyses focus on Opus, Sonnet, Astra, and Sol, the four models evaluated across the full effort sweep, and use high effort unless otherwise stated; the domain-level delegation analysis covers all nine models at max effort, except Doubao at high effort.

\subsection{Task performance}
\label{sec:main-results}

\begin{table*}[t]
\centering\begingroup\footnotesize
\setlength{\tabcolsep}{2.5pt}\renewcommand{\arraystretch}{1.28}
\setlength{\aboverulesep}{0pt}\setlength{\belowrulesep}{0pt}
\colorlet{mainoverall}{brandprimary!13}
\newcommand{\modellogo}[1]{\raisebox{-0.18em}{\includegraphics[width=1.1em,height=1.1em,keepaspectratio]{Figures/model-logos/#1.pdf}}\hspace{0.35em}}
\caption{Resolved rates across eight domains (\%). A task is resolved when its submission satisfies all acceptance criteria of its domain (Appendix~\ref{app:pass-rates}). Overall is the equal-weight mean of the domain resolved rates, and rows are sorted by Overall. Bold values denote the highest value in each column. All configurations are evaluated on the same \BenchmarkTasks{} tasks.}
\label{tab:main}
\resizebox{\textwidth}{!}{%
\begin{tabular}{@{}lcc|>{\columncolor{mainoverall}}c|*{8}{c}@{}}
\toprule
Model & Harness & Effort & \multicolumn{1}{c|}{Overall} & SWE & OPS & SCI & MATH & FIN & LAW & CDN & 3D \\
\midrule
\modellogo{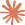}Claude Opus 5.5 & Claude Code & max & \textbf{36.95} & \textbf{29.41} & 22.22 & \textbf{56.25} & \textbf{95.83} & 12.50 & 0.00 & 57.14 & 22.22 \\
\modellogo{claude-color}Claude Opus 5.5 & Claude Code & high & 34.11 & 17.65 & 16.67 & 50.00 & 87.50 & \textbf{18.75} & 4.55 & 50.00 & \textbf{27.78} \\
\modellogo{claude-color}Claude Opus 5.5 & Claude Code & xhigh & 33.23 & 11.76 & 16.67 & 43.75 & 91.67 & 12.50 & 4.55 & 57.14 & \textbf{27.78} \\
\modellogo{claude-color}Claude Sonnet 5.5 & Claude Code & max & 33.21 & 11.76 & 11.11 & 50.00 & 79.17 & 12.50 & \textbf{9.09} & \textbf{64.29} & \textbf{27.78} \\
\modellogo{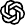}GPT-6 Astra & Codex & xhigh & 30.80 & 0.00 & 22.22 & 50.00 & 83.33 & 12.50 & 4.55 & 57.14 & 16.67 \\
\modellogo{openai}GPT-6 Astra & Codex & high & 29.82 & 0.00 & \textbf{27.78} & 43.75 & 83.33 & 12.50 & 4.55 & 50.00 & 16.67 \\
\modellogo{openai}GPT-6.1 Sol & Codex & xhigh & 29.50 & 0.00 & 16.67 & 43.75 & 79.17 & 12.50 & 4.55 & 57.14 & 22.22 \\
\modellogo{claude-color}Claude Opus 5.5 & Claude Code & medium & 29.14 & 11.76 & 16.67 & 50.00 & 70.83 & \textbf{18.75} & 0.00 & 42.86 & 22.22 \\
\modellogo{openai}GPT-6 Astra & Codex & medium & 28.11 & 0.00 & 16.67 & 43.75 & 79.17 & 12.50 & 4.55 & 57.14 & 11.11 \\
\modellogo{openai}GPT-6.1 Sol & Codex & max & 28.04 & 0.00 & 22.22 & 43.75 & 83.33 & 12.50 & 4.55 & 35.71 & 22.22 \\
\modellogo{openai}GPT-6 Astra & Codex & low & 27.80 & 5.88 & 16.67 & 43.75 & 70.83 & 12.50 & 4.55 & 57.14 & 11.11 \\
\modellogo{openai}GPT-6.1 Sol & Codex & low & 27.59 & 0.00 & 16.67 & 43.75 & 75.00 & 12.50 & 4.55 & 57.14 & 11.11 \\
\modellogo{openai}GPT-6 Astra & Codex & max & 27.34 & 0.00 & 22.22 & 43.75 & 83.33 & 12.50 & 4.55 & 35.71 & 16.67 \\
\modellogo{openai}GPT-6.1 Sol & Codex & medium & 27.30 & 0.00 & 11.11 & 43.75 & 79.17 & \textbf{18.75} & 4.55 & 50.00 & 11.11 \\
\modellogo{openai}GPT-6.1 Sol & Codex & high & 25.63 & 0.00 & 22.22 & 43.75 & 79.17 & 12.50 & 4.55 & 42.86 & 0.00 \\
\modellogo{claude-color}Claude Sonnet 5.5 & Claude Code & high & 25.26 & 0.00 & 11.11 & 43.75 & 62.50 & 12.50 & 0.00 & 50.00 & 22.22 \\
\modellogo{claude-color}Claude Sonnet 5.5 & Claude Code & xhigh & 23.24 & 0.00 & 11.11 & 43.75 & 70.83 & 6.25 & 0.00 & 42.86 & 11.11 \\
\modellogo{claude-color}Claude Opus 5.5 & Claude Code & low & 22.02 & 0.00 & 11.11 & 43.75 & 66.67 & 6.25 & 0.00 & 42.86 & 5.56 \\
\modellogo{claude-color}Claude Sonnet 5.5 & Claude Code & medium & 19.78 & 0.00 & 11.11 & 37.50 & 58.33 & \textbf{18.75} & 0.00 & 21.43 & 11.11 \\
\modellogo{claude-color}Claude Sonnet 5.5 & Claude Code & low & 16.88 & 0.00 & 5.56 & 37.50 & 54.17 & 6.25 & 4.55 & 21.43 & 5.56 \\
\modellogo{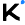}Kimi K3.1 Pro & Claude Code & max & 16.48 & 0.00 & 16.67 & 43.75 & 37.50 & 12.50 & 0.00 & 21.43 & 0.00 \\
\modellogo{openai}GPT-6 Luna & Codex & max & 13.32 & 0.00 & 5.56 & 37.50 & 25.00 & 12.50 & 4.55 & 21.43 & 0.00 \\
\modellogo{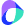}Doubao-Seed-Evolving & Claude Code & high & 10.84 & 0.00 & \textbf{27.78} & 31.25 & 0.00 & 6.25 & 0.00 & 21.43 & 0.00 \\
\modellogo{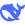}DeepSeek V4.1 Flash & Claude Code & max & 6.08 & 0.00 & 11.11 & 25.00 & 0.00 & 12.50 & 0.00 & 0.00 & 0.00 \\
\modellogo{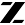}GLM-5.3 Flash & Claude Code & max & 4.43 & 0.00 & 16.67 & 18.75 & 0.00 & 0.00 & 0.00 & 0.00 & 0.00 \\
\bottomrule
\end{tabular}}\endgroup
\end{table*}

\paragraph{Resolved rates reveal substantial headroom.}
Table~\ref{tab:main} reports, for each configuration, the share of tasks whose submissions satisfy every acceptance criterion of their domain. The best configuration, Opus at max effort, resolves 36.95\% of tasks on average across domains, followed by Opus at high effort with 34.11\%. Only MATH approaches saturation, with a best resolved rate of 95.83\%; the highest SWE resolved rate is 29.41\%, and LAW and FIN resolved rates never exceed 9.09\% and 18.75\%. No single configuration leads every domain. Opus at max effort leads SWE, SCI, and MATH, Sonnet at max effort leads LAW and CDN and shares the lead in 3D, and OPS is led by Astra at high effort and by Doubao, both at 27.78\%. Formal mathematics separates the models most sharply: the strongest configurations resolve more than 90\% of MATH tasks, Kimi and Luna resolve 37.50\% and 25.00\%, and Doubao, GLM, and DeepSeek resolve none. Appendix~\ref{app:score-resources} reports continuous scores, which credit partial progress and are correspondingly higher.

\paragraph{Effort matters far more for the Claude models.}
Both Claude models resolve the most tasks at max effort and both GPT models at xhigh. Across the five settings, however, Opus ranges from 22.02\% to 36.95\% and Sonnet from 16.88\% to 33.21\%, whereas Astra varies only between 27.34\% and 30.80\% and Sol between 25.63\% and 29.50\%. Higher effort is not uniformly beneficial: Sol at low effort resolves more tasks than at medium or high effort.

\subsection{Resource use}
\label{sec:resources}

\begin{figure}[t]
\centering
\includegraphics[width=.98\linewidth]{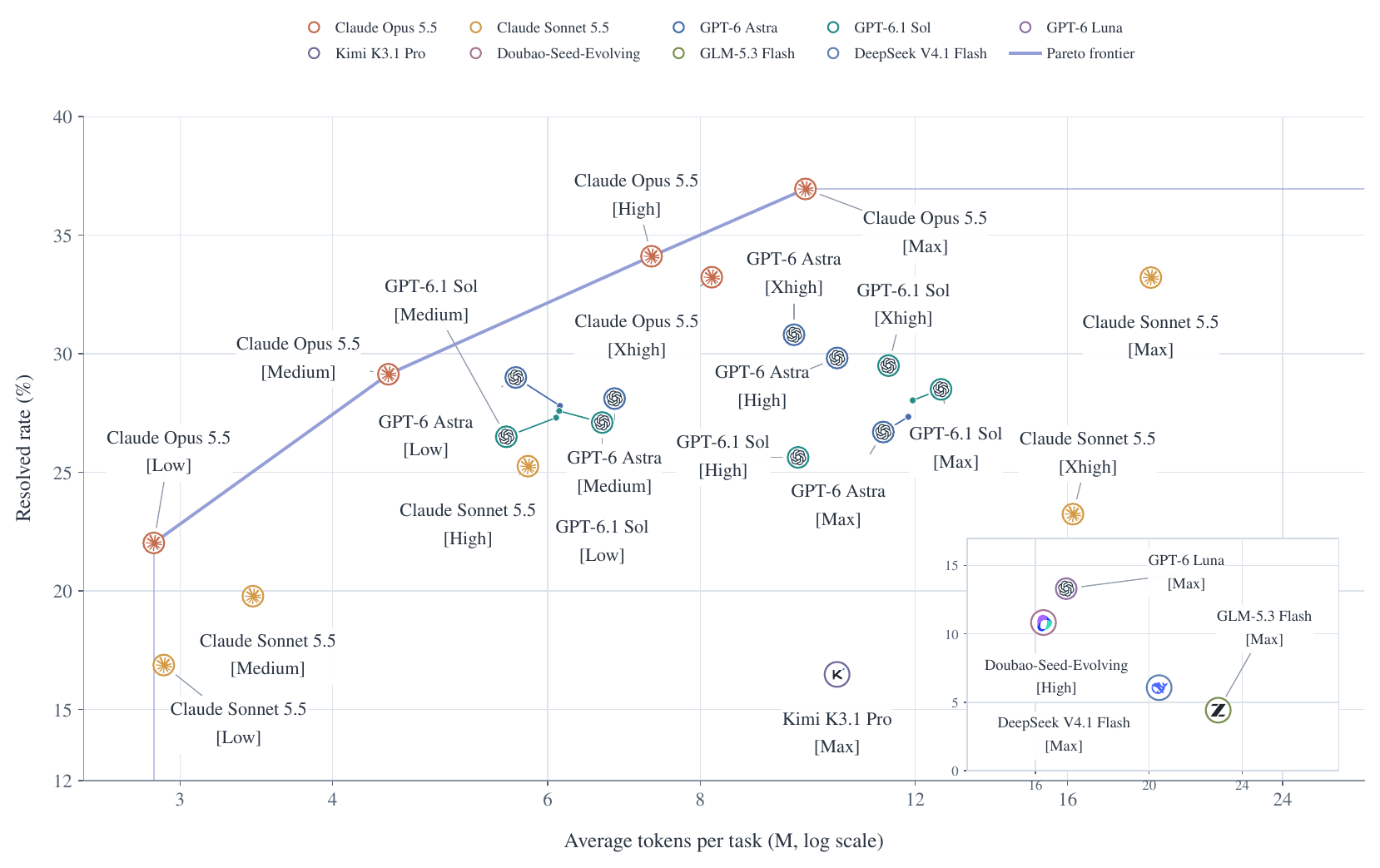}
\caption{\textbf{Overall resolved rate versus token consumption.} Resolved rates match Table~\ref{tab:main}. Mean per-task tokens include main-agent and worker usage. The token axis is logarithmic, and the inset uses its own axes. The line connects non-dominated configurations.}
\label{fig:resolved-tokens}
\end{figure}

\begin{figure}[t]
\centering
\includegraphics[width=.98\linewidth]{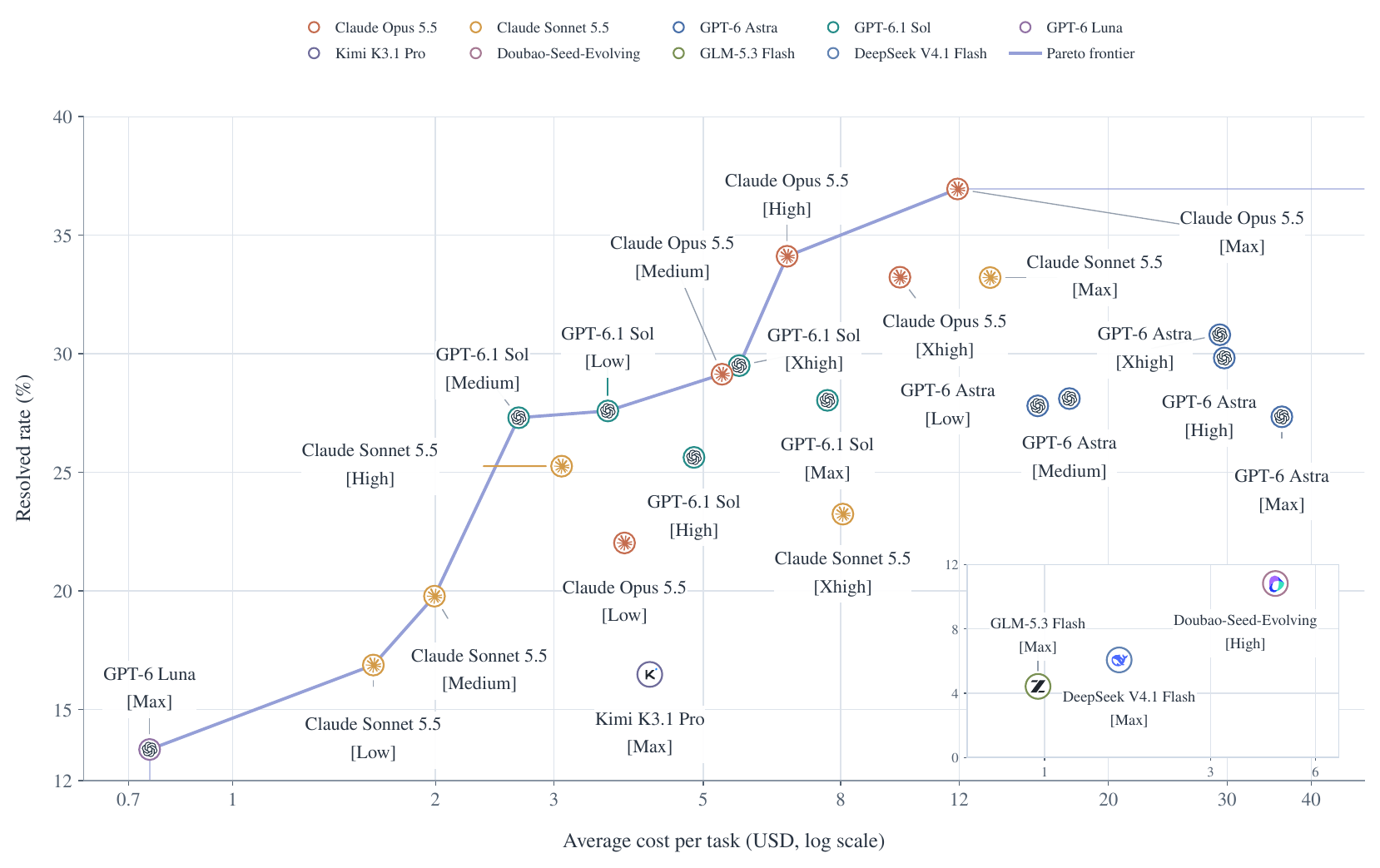}
\caption{\textbf{Overall resolved rate versus mean inference cost per task.} Resolved rates match Table~\ref{tab:main}. Costs include the main agent and workers and are estimated from recorded token usage and model-specific API prices. The cost axis is logarithmic, and the line connects non-dominated configurations.}
\label{fig:resolved-cost}
\end{figure}

\paragraph{Returns on additional tokens vary across systems.}
Figure~\ref{fig:resolved-tokens} pairs each configuration's Overall resolved rate with its mean token consumption. The token frontier consists entirely of Opus configurations, and Opus at high effort resolves 34.11\% of tasks with 7.36M tokens per task, a rate that no other model reaches at any token budget. Sonnet at max effort uses 3.2$\times$ the tokens of its high-effort configuration and gains 7.95 points of resolved rate, whereas Sol at xhigh effort uses 1.9$\times$ the tokens of its low-effort configuration for a gain of only 1.91 points.

\paragraph{Pricing reorders the resource comparison.}
Figure~\ref{fig:resolved-cost} replaces token volume with estimated inference cost. Because per-token prices differ by more than an order of magnitude, the cost frontier is far more mixed than the token frontier, running from Luna through low-effort Sonnet and several Sol configurations to Opus. Sol at low effort resolves 27.59\% of tasks at \$3.61 per task and Opus at high effort 34.11\% at \$6.66, whereas Astra at xhigh effort resolves 30.80\% at \$29.26, 3.31 points below Opus at high effort at 4.4$\times$ its cost.

\subsection{Delegation patterns}
\label{sec:workflow-usage}

We characterize delegation with two quantities. \emph{Workflow calls per task} is the total number of Workflow invocations divided by the number of tasks, including tasks without invocations. \emph{Worker invocations per Workflow} is the total number of worker invocations divided by the number of launched Workflows. A Workflow invocation is an invocation-start event deduplicated by call identifier; list, query, and read operations are excluded.

\begin{figure}[t]
\centering
\includegraphics[width=.94\linewidth]{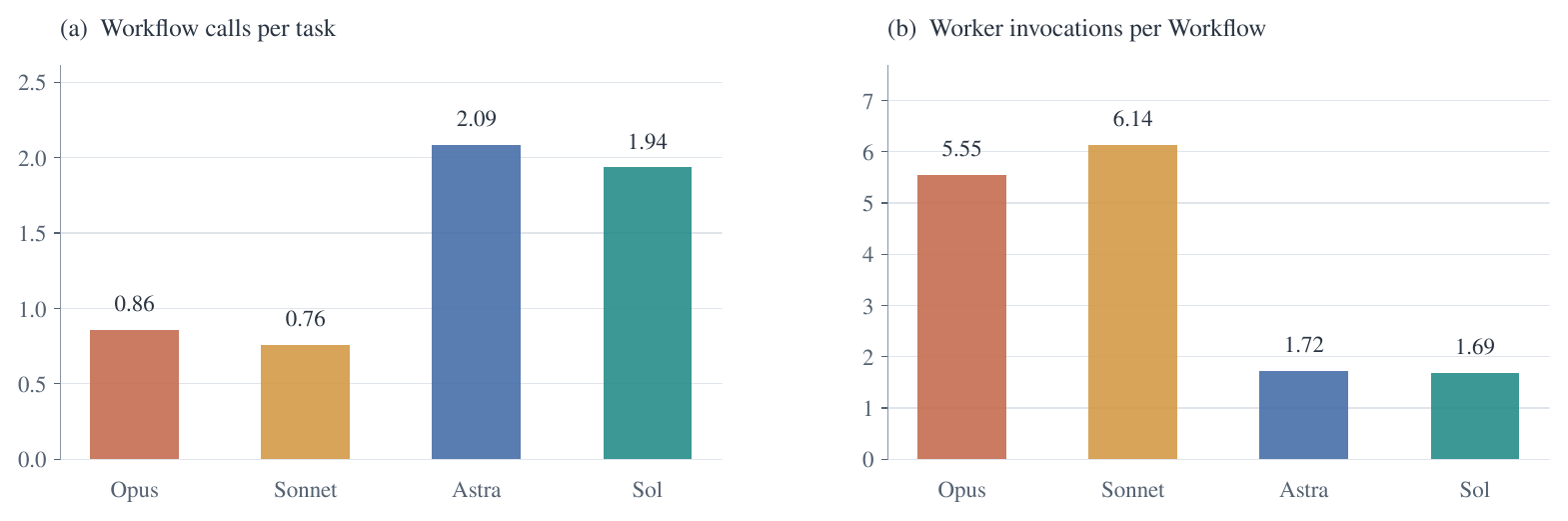}
\caption{\textbf{Invocation frequency and delegation size at high effort.} Mean calls per task include executions with no calls; worker invocations per Workflow uses launched Workflows as its denominator.}
\label{fig:workflow-frequency-size}
\end{figure}

\begin{figure}[t]
\centering
\includegraphics[width=.96\linewidth]{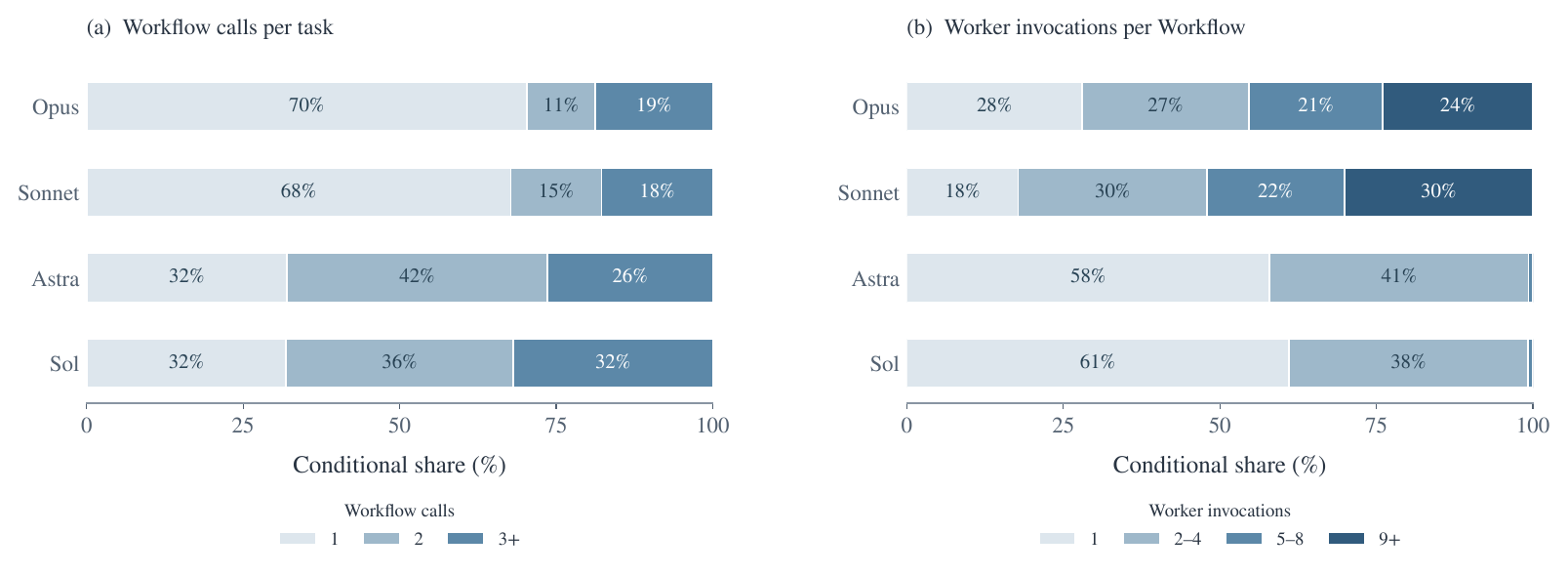}
\caption{\textbf{Distribution of delegation frequency and size at high effort.} Panel (a) shows Workflow calls per task, and panel (b) shows worker invocations per Workflow. Distributions are normalized over tasks with at least one Workflow call and Workflows with at least one worker invocation, respectively.}
\label{fig:workflow-size-distribution}
\end{figure}

\paragraph{Claude Code systems delegate less often but in larger groups.}
At high effort, Opus and Sonnet in Claude Code average 0.86 and 0.76 Workflow calls per task with 5.55 and 6.14 worker invocations per Workflow, whereas Astra and Sol in Codex average 2.09 and 1.94 calls with only 1.72 and 1.69 worker invocations (Figure~\ref{fig:workflow-frequency-size}). The difference holds across the distribution (Figure~\ref{fig:workflow-size-distribution}). Among tasks with at least one Workflow call, a single call accounts for 70\% of Opus and 68\% of Sonnet executions, compared with about 32\% for Astra and Sol; among nonempty Workflows, groups of nine or more workers make up 24\% of Opus and 30\% of Sonnet invocations, whereas about 99\% of Astra and Sol Workflows contain at most four workers.

\begin{figure}[t]
\centering
\includegraphics[width=.94\linewidth]{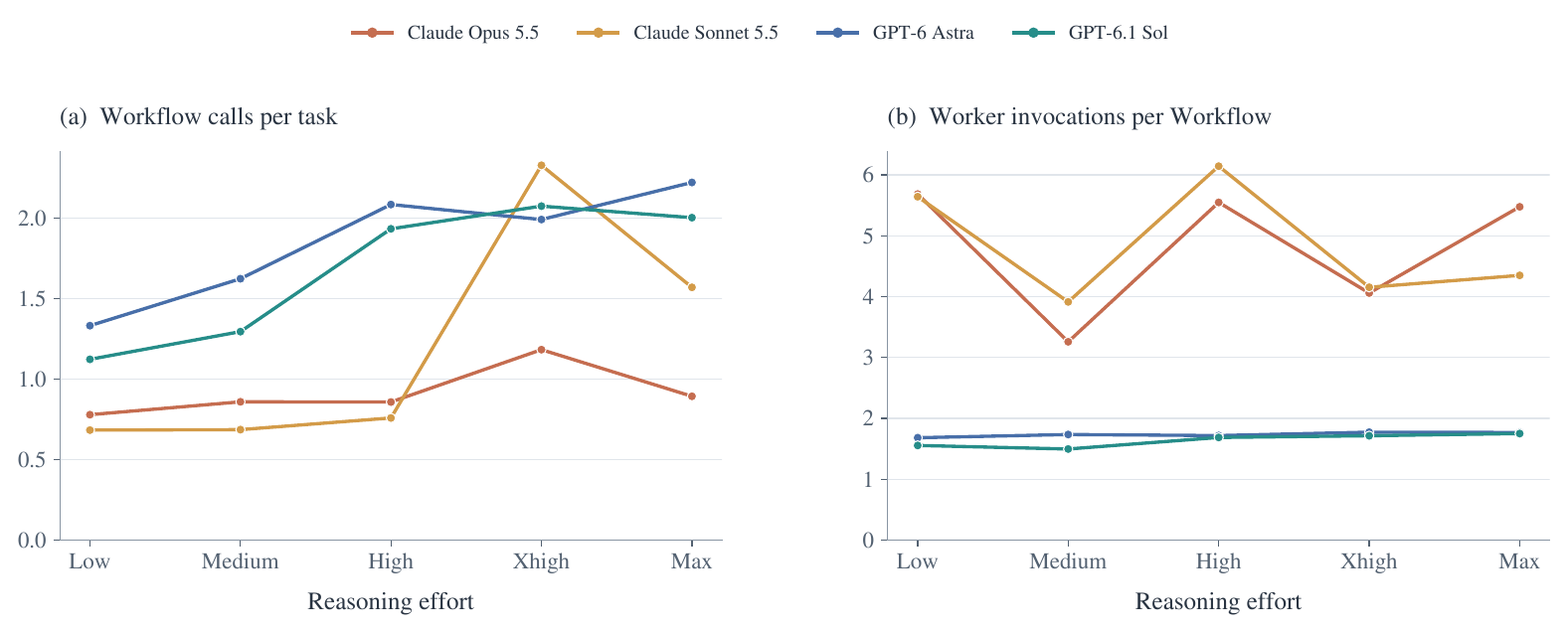}
\caption{\textbf{Delegation across reasoning efforts.} Panel (a) reports Workflow calls per task, including zero-call executions; panel (b) reports worker invocations per Workflow. Lines connect effort settings within each system.}
\label{fig:workflow-effort}
\end{figure}

Effort shifts delegation differently in the two harness families (Figure~\ref{fig:workflow-effort}). From low to max effort, Astra's calls per task rise by 67\%, from 1.33 to 2.22, while its worker invocations per Workflow change by only 5\%; Sol's calls rise by 79\%, from 1.12 to 2.00, with a 12\% change in worker invocations. Both Claude Code systems instead reach their highest call frequency at xhigh, and Sonnet moves from 2.33 calls and 4.15 worker invocations per Workflow at xhigh to 1.57 calls and approximately 4.35 worker invocations at max.

\begin{figure}[t]
\centering
\includegraphics[width=\linewidth]{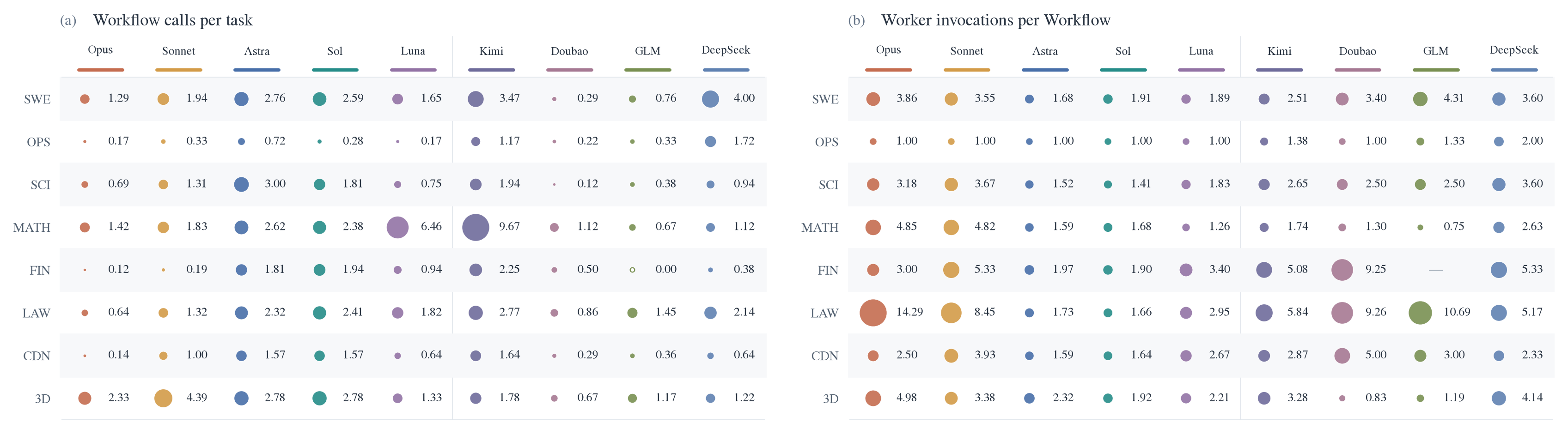}
\caption{\textbf{Delegation patterns across eight domains and nine models.} All models use max effort except Doubao-Seed-Evolving, which uses high. Left: total Workflow calls divided by the number of tasks in each domain, including tasks with no calls. Right: total worker invocations divided by the number of Workflows. Circle area encodes value within each panel, with means rounded to two decimal places printed alongside. Colors identify models; a dash indicates no Workflows.}
\label{fig:domain-workflow-patterns}
\end{figure}

\paragraph{The domain shapes delegation within each system.}
Figure~\ref{fig:domain-workflow-patterns} breaks both quantities down by domain at max effort. In LAW, Opus and Sonnet average 14.29 and 8.45 worker invocations per Workflow against 1.73 for Astra and 1.66 for Sol, while their call frequencies show the reverse ordering. Frequent delegation can also consist of small Workflows: Kimi and Luna make 9.67 and 6.46 calls per task in MATH but average only 1.74 and 1.26 worker invocations per Workflow, although Kimi's groups grow to 5.84 and 5.08 workers in LAW and FIN. OPS delegations remain small for every model, at 0.17--1.72 calls per task and 1.00--2.00 worker invocations per Workflow. Call frequency alone therefore does not characterize the scale of collaboration.

\subsection{Workflow transfer across execution models}
\label{sec:workflow-transfer}

The preceding analyses characterize how systems organize their work, but system-level outcomes combine the quality of a workflow with the capability of the agents that carry it out. To separate the two, we hold the executor fixed and vary only the source of the workflow on a challenging subset of 32 tasks (Appendix~\ref{app:transfer}). Astra and Opus serve as workflow sources at max effort, and Kimi, DeepSeek, Luna, and Doubao serve as execution models; every worker is an instance of the execution model. In the \emph{static} mode, the source model reads the task instruction and the clean initial workspace, writes a Workflow program, and then stops; the frozen program runs in a fresh workspace, where its internal branches, loops, and parallel stages still execute. In the \emph{dynamic} mode, which follows Claude Code dynamic workflows~\citep{anthropic2026dynamicworkflows}, the source model launches Workflows, observes worker results, and revises the workflow throughout execution. Unlike the main agent in \S\ref{sec:protocol}, the source model cannot edit the workspace in either mode, so all task work is performed by the execution model's workers. Workflow generation counts against the same wall-clock budget. For each execution model, we compare the four transferred settings against its own dynamic workflow (Table~\ref{tab:workflow-transfer}). Because resolved rates on a 32-task subset change in coarse steps, we report continuous scores for this experiment to capture finer-grained differences between settings.

\begin{table}[htbp]
\centering
\caption{\textbf{Workflow transfer on a challenging subset.} Astra and Opus generate workflows at max effort for execution by the models listed below. Self denotes the execution model acting as its own source in the dynamic mode. Bold and underlined values indicate the best and second-best scores in each row.}
\label{tab:workflow-transfer}
\begingroup
\small
\setlength{\tabcolsep}{7pt}
\renewcommand{\arraystretch}{1.30}
\providecommand{\transferlogoroot}{Figures/model-logos}
\newcommand{\transferlogo}[1]{\raisebox{-0.18em}{\includegraphics[width=1.05em,height=1.05em,keepaspectratio]{\transferlogoroot/#1.pdf}}\hspace{0.4em}}
\begin{tabular*}{\linewidth}{@{\extracolsep{\fill}}lccccc@{}}
\toprule
& Self & \multicolumn{2}{c}{\transferlogo{openai}GPT-6 Astra} & \multicolumn{2}{c}{\transferlogo{claude-color}Claude Opus 5.5} \\
\cmidrule(lr){2-2}\cmidrule(lr){3-4}\cmidrule(l){5-6}
Execution model & Dynamic & Static & Dynamic & Static & Dynamic \\
\midrule
\transferlogo{kimi-color}Kimi K3.1 Pro & 35.75 & 29.69 & 33.80 & \underline{45.04} & \textbf{46.61} \\
\transferlogo{openai}GPT-6 Luna & 25.08 & 24.87 & \underline{29.22} & \textbf{30.37} & 25.35 \\
\transferlogo{doubao-color}Doubao-Seed-Evolving & 20.39 & 22.07 & 24.25 & \textbf{28.04} & \underline{27.46} \\
\transferlogo{deepseek-color}DeepSeek V4.1 Flash & 18.37 & 12.85 & 17.69 & \underline{21.79} & \textbf{25.24} \\
\bottomrule
\end{tabular*}
\endgroup
\end{table}

\paragraph{Workflows from Opus improve every executor.}
With Opus dynamic workflows, Kimi improves from 35.75 to 46.61 and DeepSeek from 18.37 to 25.24, while Luna and Doubao obtain their highest scores with Opus static workflows, at 30.37 and 28.04. Averaged over the four executors, Opus workflows add 6.41 points in the static mode and 6.27 in the dynamic mode. Because the source model never edits the workspace, these gains arise from the workflow it authors, including the instructions it gives each worker, rather than from work performed by the source.

\paragraph{Orchestration quality differs from task performance.}
Astra workflows transfer far less well, averaging $-2.53$ points in the static mode and $+1.34$ in the dynamic mode; they lower the scores of Kimi and DeepSeek in both modes and exceed the executor's own workflow only for Doubao in both modes and for Luna in the dynamic mode. Yet Astra is itself a far stronger system than any of the four executors, resolving 27.34\% of tasks at max effort compared with at most 16.48\% for the executors. Strong solo performance therefore does not by itself yield workflows that help other models, which suggests that the ability to orchestrate others is not simply a byproduct of the ability to complete the task.

\subsection{Case study: orchestrating 3D scene construction}
\label{sec:case-study}

\begin{figure*}[t]
    \centering
    \includegraphics[width=\textwidth]{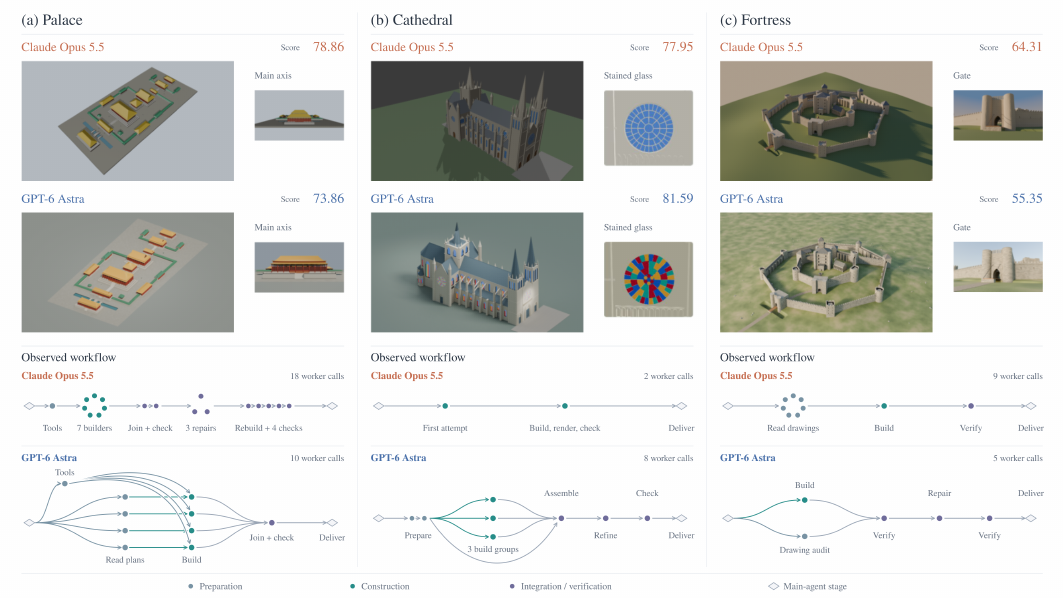}
    \caption{\textbf{Artifacts and workflows for three 3D construction tasks.} Each column pairs Opus in Claude Code with Astra in Codex. Upper panels show overview renders, scores, and architectural details; lower panels reconstruct the workflows and their worker calls. Dot colors follow the legend and mark preparation or drawing analysis, initial construction, and integration, checking, repair, or rebuilding. Dots represent worker calls and diamonds main-agent stages; arrows indicate dependencies and handoffs.}
    \label{fig:3d_case_studies}
\end{figure*}

The aggregate measures in \S\ref{sec:workflow-usage} summarize how often and how widely systems delegate, but not what the resulting workflows look like. Three 3D construction tasks illustrate distinct forms of workflow organization: iterative component integration in the palace, spatial decomposition in the cathedral, and drawing analysis for centralized construction in the fortress. Figure~\ref{fig:3d_case_studies} pairs the delivered artifacts with the corresponding workflows of Opus in Claude Code and Astra in Codex. Each task requires a Blender scene, renders, and a manifest that reconciles construction drawings with a change order.

In the palace, both workflows distribute construction across components: Opus runs seven builders followed by assembly and iterative repairs, whereas Astra first establishes a shared foundation and then assigns assembly to a single worker. In the cathedral, Opus makes two sequential whole-scene attempts, whereas Astra divides the scene among three regional builders and a separate assembly worker, which introduces interfaces at shared structural positions. In the fortress, both workflows rely on a single whole-scene builder, but Opus precedes construction with seven drawing-analysis calls, whereas Astra audits the drawings concurrently with construction. Delegation therefore serves two distinct purposes across these tasks, building components in parallel and preparing the inputs of a central builder, and integration and repair enter the workflow at different stages.

\section{Conclusion}

WorkflowArena evaluates agents that organize their own work on long-horizon tasks. Its \BenchmarkTasks{} tasks span \BenchmarkDomains{} domains and \BenchmarkForms{} task forms, and each pairs professional source material or an authored scenario with an explicit initial state, an interdependent deliverable, and hidden domain-specific verification. Every delegation passes through a recorded interface under a shared time budget, so each result can be read together with the workflow that produced it.

Returning to the question posed in the introduction, current agents can orchestrate multi-agent workflows for long-horizon tasks, but they remain far from accomplishing them reliably: the strongest configuration, Opus with Claude Code at max effort, fully resolves only 36.95\% of tasks on average across domains, and no configuration leads every domain. How the agents organize their work also differs systematically. At high effort, the Claude models in Claude Code delegate less often but in larger groups, whereas the GPT models in Codex delegate more often in smaller units, and within each system the task domain further shapes how often and how widely work is delegated. Holding the executor fixed on a challenging subset shows that this organization matters in its own right: workflows authored by Opus raise the score of every execution model, while workflows from Astra, itself a much stronger system than these executors, help far less. The quality of orchestration thus appears to be a measurable source of performance that is distinct from the ability to complete the task alone.

\paragraph{Limitations.}
Each configuration is evaluated once per task, so differences between closely ranked configurations are reported without variance estimates. Each model is evaluated in a single harness, and differences between the Claude and GPT systems therefore reflect the model and its harness jointly; the non-GPT models that run in Claude Code provide only a partial basis for separating the two. Content and 3D deliverables are graded against task rubrics by an LLM judge, and agreement with human graders is not measured in this work. The transfer experiment uses a subset of the benchmark and two source models, and because source models also write the instructions given to workers, its gains may combine better organization with task knowledge conveyed in those instructions.

Because WorkflowArena records both the deliverable and the workflow that produced it, it also supports future study of heterogeneous systems in which the orchestrator and its workers use different models, as well as adaptive budgets and interactive task execution.

\par\FloatBarrier

\section*{Acknowledgements}
This project is funded and supported by Navers Lab, Einsia.AI.

\bibliographystyle{unsrtnat}
\begingroup
\renewcommand{\bibfont}{\fontsize{8.5}{10}\selectfont}
\setlength{\bibsep}{1pt plus 0.5pt}
\bibliography{ref}

\begin{thebibliography}{42}
\providecommand{\natexlab}[1]{#1}
\providecommand{\url}[1]{\texttt{#1}}
\expandafter\ifx\csname urlstyle\endcsname\relax
  \providecommand{\doi}[1]{doi: #1}\else
  \providecommand{\doi}{doi: \begingroup \urlstyle{rm}\Url}\fi

\bibitem[Yao et~al.(2022)Yao, Zhao, Yu, Du, Shafran, Narasimhan, and
  Cao]{yao2022react}
Shunyu Yao, Jeffrey Zhao, Dian Yu, Nan Du, Izhak Shafran, Karthik Narasimhan,
  and Yuan Cao.
\newblock {ReAct: Synergizing Reasoning and Acting in Language Models}.
\newblock \emph{arXiv preprint arXiv:2210.03629}, 2022.
\newblock URL \url{https://arxiv.org/abs/2210.03629}.

\bibitem[Shinn et~al.(2023)Shinn, Cassano, Berman, Gopinath, Narasimhan, and
  Yao]{shinn2023reflexion}
Noah Shinn, Federico Cassano, Edward Berman, Ashwin Gopinath, Karthik
  Narasimhan, and Shunyu Yao.
\newblock {Reflexion: Language Agents with Verbal Reinforcement Learning}.
\newblock \emph{arXiv preprint arXiv:2303.11366}, 2023.
\newblock URL \url{https://arxiv.org/abs/2303.11366}.

\bibitem[Yang et~al.(2024)Yang, Jimenez, Wettig, Lieret, Yao, Narasimhan, and
  Press]{yang2024sweagent}
John Yang, Carlos~E. Jimenez, Alexander Wettig, Kilian Lieret, Shunyu Yao,
  Karthik Narasimhan, and Ofir Press.
\newblock {SWE-agent: Agent-Computer Interfaces Enable Automated Software
  Engineering}.
\newblock \emph{arXiv preprint arXiv:2405.15793}, 2024.
\newblock URL \url{https://arxiv.org/abs/2405.15793}.

\bibitem[Feng et~al.(2026)Feng, Xiang, Yang, Ma, Chen, Zhang, Huang, Wu, Liu,
  Wang, He, Wang, Hong, Chen, Bei, Wang, Chen, Zhang, Dai, Luo, Pan, Wang, Fan,
  Wang, Zou, Song, Wang, Li, Hu, Li, Huang, Zhang, Su, Zhang, and
  Chang]{feng2026graphengineering}
Yuyuan Feng, Zhishang Xiang, Chaobin Yang, Qichao Ma, Zerui Chen, Yujing Zhang,
  Ke~Huang, Chuanjie Wu, Zhaoxu Liu, Yili Wang, Xin He, Jiapu Wang, Zijin Hong,
  Hao Chen, Yuanchen Bei, Kun Wang, Shengyuan Chen, Ningyu Zhang, Enyan Dai,
  Linhao Luo, Qingyi Pan, Qi~Wang, Wenqi Fan, Guangjing Wang, Na~Zou, Yangqiu
  Song, Xin Wang, Zechao Li, Xia Hu, Qing Li, Xiao Huang, Zhihong Zhang,
  Jinsong Su, Qinggang Zhang, and Yi~Chang.
\newblock {Graph Engineering in the Era of LLM Agents: From Individual
  Intelligence to System Intelligence}.
\newblock \emph{arXiv preprint arXiv:2608.21156}, 2026.
\newblock URL \url{https://arxiv.org/abs/2608.21156}.

\bibitem[Fourney et~al.(2024)Fourney, Bansal, Mozannar, Tan, Salinas, Zhu,
  Niedtner, Proebsting, Bassman, Gerrits, Alber, Chang, Loynd, West, Dibia,
  Awadallah, Kamar, Hosn, and Amershi]{fourney2024magenticone}
Adam Fourney, Gagan Bansal, Hussein Mozannar, Cheng Tan, Eduardo Salinas,
  Erkang Zhu, Friederike Niedtner, Grace Proebsting, Griffin Bassman, Jack
  Gerrits, Jacob Alber, Peter Chang, Ricky Loynd, Robert West, Victor Dibia,
  Ahmed Awadallah, Ece Kamar, Rafah Hosn, and Saleema Amershi.
\newblock {Magentic-One: A Generalist Multi-Agent System for Solving Complex
  Tasks}.
\newblock \emph{arXiv preprint arXiv:2411.04468}, 2024.
\newblock URL \url{https://arxiv.org/abs/2411.04468}.

\bibitem[Shihipar and Bidasaria(2026)]{anthropic2026dynamicworkflows}
Thariq Shihipar and Sid Bidasaria.
\newblock A harness for every task: Dynamic workflows in {Claude Code}.
\newblock Claude Developer Blog, 2026.
\newblock URL
  \url{https://claude.dev/blog/a-harness-for-every-task-dynamic-workflows-in-claude-code/}.

\bibitem[Liu et~al.(2026)Liu, Li, Wang, and Wang]{liu2026dynaflow}
Ruhao Liu, Qi~Li, Shizun Wang, and Xinchao Wang.
\newblock {DynaFlow: Learning Free-Form Multi-Agent Orchestration}.
\newblock \emph{arXiv preprint arXiv:2610.34467}, 2026.
\newblock URL \url{https://arxiv.org/abs/2610.34467}.

\bibitem[{Anthropic}(2024)]{anthropic2024effectiveagents}
{Anthropic}.
\newblock Building effective agents.
\newblock Anthropic Engineering Blog, 2024.
\newblock URL
  \url{https://www.anthropic.com/research/building-effective-agents}.

\bibitem[Kim et~al.(2023)Kim, Moon, Tabrizi, Lee, Mahoney, Keutzer, and
  Gholami]{kim2023llmcompiler}
Sehoon Kim, Suhong Moon, Ryan Tabrizi, Nicholas Lee, Michael~W. Mahoney, Kurt
  Keutzer, and Amir Gholami.
\newblock {An LLM Compiler for Parallel Function Calling}.
\newblock \emph{arXiv preprint arXiv:2312.04511}, 2023.
\newblock URL \url{https://arxiv.org/abs/2312.04511}.

\bibitem[Liu et~al.(2023{\natexlab{a}})Liu, Yu, Zhang, Xu, Lei, Lai, Gu, Ding,
  Men, Yang, Zhang, Deng, Zeng, Du, Zhang, Shen, Zhang, Su, Sun, Huang, Dong,
  and Tang]{liu2023agentbench}
Xiao Liu, Hao Yu, Hanchen Zhang, Yifan Xu, Xuanyu Lei, Hanyu Lai, Yu~Gu,
  Hangliang Ding, Kaiwen Men, Kejuan Yang, Shudan Zhang, Xiang Deng, Aohan
  Zeng, Zhengxiao Du, Chenhui Zhang, Sheng Shen, Tianjun Zhang, Yu~Su, Huan
  Sun, Minlie Huang, Yuxiao Dong, and Jie Tang.
\newblock {AgentBench: Evaluating LLMs as Agents}.
\newblock \emph{arXiv preprint arXiv:2308.03688}, 2023{\natexlab{a}}.
\newblock URL \url{https://arxiv.org/abs/2308.03688}.

\bibitem[Zhou et~al.(2023)Zhou, Xu, Zhu, Zhou, Lo, Sridhar, Cheng, Ou, Bisk,
  Fried, Alon, and Neubig]{zhou2024webarena}
Shuyan Zhou, Frank~F. Xu, Hao Zhu, Xuhui Zhou, Robert Lo, Abishek Sridhar,
  Xianyi Cheng, Tianyue Ou, Yonatan Bisk, Daniel Fried, Uri Alon, and Graham
  Neubig.
\newblock {WebArena: A Realistic Web Environment for Building Autonomous
  Agents}.
\newblock \emph{arXiv preprint arXiv:2307.13854}, 2023.
\newblock URL \url{https://arxiv.org/abs/2307.13854}.

\bibitem[Xie et~al.(2024)Xie, Zhang, Chen, Li, Zhao, Cao, Hua, Cheng, Shin,
  Lei, Liu, Xu, Zhou, Savarese, Xiong, Zhong, and Yu]{xie2024osworld}
Tianbao Xie, Danyang Zhang, Jixuan Chen, Xiaochuan Li, Siheng Zhao, Ruisheng
  Cao, Toh~Jing Hua, Zhoujun Cheng, Dongchan Shin, Fangyu Lei, Yitao Liu,
  Yiheng Xu, Shuyan Zhou, Silvio Savarese, Caiming Xiong, Victor Zhong, and Tao
  Yu.
\newblock {OSWorld: Benchmarking Multimodal Agents for Open-Ended Tasks in Real
  Computer Environments}.
\newblock \emph{arXiv preprint arXiv:2404.07972}, 2024.
\newblock URL \url{https://arxiv.org/abs/2404.07972}.

\bibitem[Trivedi et~al.(2024)Trivedi, Khot, Hartmann, Manku, Dong, Li, Gupta,
  Sabharwal, and Balasubramanian]{trivedi2024appworld}
Harsh Trivedi, Tushar Khot, Mareike Hartmann, Ruskin Manku, Vinty Dong, Edward
  Li, Shashank Gupta, Ashish Sabharwal, and Niranjan Balasubramanian.
\newblock {AppWorld: A Controllable World of Apps and People for Benchmarking
  Interactive Coding Agents}.
\newblock \emph{arXiv preprint arXiv:2407.18901}, 2024.
\newblock URL \url{https://arxiv.org/abs/2407.18901}.

\bibitem[Jimenez et~al.(2023)Jimenez, Yang, Wettig, Yao, Pei, Press, and
  Narasimhan]{jimenez2024swebench}
Carlos~E. Jimenez, John Yang, Alexander Wettig, Shunyu Yao, Kexin Pei, Ofir
  Press, and Karthik Narasimhan.
\newblock {SWE-bench: Can Language Models Resolve Real-World GitHub Issues?}
\newblock \emph{arXiv preprint arXiv:2310.06770}, 2023.
\newblock URL \url{https://arxiv.org/abs/2310.06770}.

\bibitem[Boisvert et~al.(2024)Boisvert, Thakkar, Gasse, Caccia, De~Chezelles,
  Cappart, Chapados, Lacoste, and Drouin]{boisvert2024workarenapp}
L{\'{e}}o Boisvert, Megh Thakkar, Maxime Gasse, Massimo Caccia, Thibault
  Le~Sellier De~Chezelles, Quentin Cappart, Nicolas Chapados, Alexandre
  Lacoste, and Alexandre Drouin.
\newblock {WorkArena++: Towards Compositional Planning and Reasoning-based
  Common Knowledge Work Tasks}.
\newblock \emph{arXiv preprint arXiv:2407.05291}, 2024.
\newblock URL \url{https://arxiv.org/abs/2407.05291}.

\bibitem[Zhu et~al.(2025)Zhu, Du, Hong, Yang, Guo, Wang, Wang, Qian, Tang, Ji,
  and You]{zhu2025multiagentbench}
Kunlun Zhu, Hongyi Du, Zhaochen Hong, Xiaocheng Yang, Shuyi Guo, Zhe Wang,
  Zhenhailong Wang, Cheng Qian, Xiangru Tang, Heng Ji, and Jiaxuan You.
\newblock {MultiAgentBench: Evaluating the Collaboration and Competition of LLM
  agents}.
\newblock \emph{arXiv preprint arXiv:2503.01935}, 2025.
\newblock URL \url{https://arxiv.org/abs/2503.01935}.

\bibitem[Mediratta et~al.(2026)Mediratta, Xiao, Stone, Biswas, and
  Stengel-Eskin]{mediratta2026parallact}
Rishabh Mediratta, Lingyun Xiao, Peter Stone, Joydeep Biswas, and Elias
  Stengel-Eskin.
\newblock {Decentralized Multi-Agent Coordination and Communication in Shared
  Codebases}.
\newblock \emph{arXiv preprint arXiv:2611.00314}, 2026.
\newblock URL \url{https://arxiv.org/abs/2611.00314}.

\bibitem[Xiao et~al.(2024)Xiao, Ma, Wang, Wu, Zhao, Wang, Huang, and
  Li]{xiao2024flowbench}
Ruixuan Xiao, Wentao Ma, Ke~Wang, Yuchuan Wu, Junbo Zhao, Haobo Wang, Fei
  Huang, and Yongbin Li.
\newblock {FlowBench: Revisiting and Benchmarking Workflow-Guided Planning for
  LLM-based Agents}.
\newblock \emph{arXiv preprint arXiv:2406.14884}, 2024.
\newblock URL \url{https://arxiv.org/abs/2406.14884}.

\bibitem[Qiao et~al.(2024)Qiao, Fang, Qiu, Wang, Zhang, Jiang, Xie, Huang, and
  Chen]{qiao2025worfbench}
Shuofei Qiao, Runnan Fang, Zhisong Qiu, Xiaobin Wang, Ningyu Zhang, Yong Jiang,
  Pengjun Xie, Fei Huang, and Huajun Chen.
\newblock {Benchmarking Agentic Workflow Generation}.
\newblock \emph{arXiv preprint arXiv:2410.07869}, 2024.
\newblock URL \url{https://arxiv.org/abs/2410.07869}.

\bibitem[Ren et~al.(2026)Ren, He, Zhang, Qian, Han, Zheng, Li, and
  Zhang]{ren2026orchbench}
Zhenzhen Ren, Jiyan He, Xinpeng Zhang, Zhenxing Qian, Ke~Han, Shuxin Zheng,
  GuoBiao Li, and Xiaoqing Zhang.
\newblock {OrchBench: Evaluating Multi-Agent Orchestration Plans in Isolation
  via Deterministic Simulation}.
\newblock \emph{arXiv preprint arXiv:2607.25656}, 2026.
\newblock URL \url{https://arxiv.org/abs/2607.25656}.

\bibitem[Mialon et~al.(2023)Mialon, Fourrier, Swift, Wolf, LeCun, and
  Scialom]{mialon2023gaia}
Gr{\'{e}}goire Mialon, Cl{\'{e}}mentine Fourrier, Craig Swift, Thomas Wolf,
  Yann LeCun, and Thomas Scialom.
\newblock {GAIA: a benchmark for General AI Assistants}.
\newblock \emph{arXiv preprint arXiv:2311.12983}, 2023.
\newblock URL \url{https://arxiv.org/abs/2311.12983}.

\bibitem[Yoran et~al.(2024)Yoran, Amouyal, Malaviya, Bogin, Press, and
  Berant]{yoran2024assistantbench}
Ori Yoran, Samuel~Joseph Amouyal, Chaitanya Malaviya, Ben Bogin, Ofir Press,
  and Jonathan Berant.
\newblock {AssistantBench: Can Web Agents Solve Realistic and Time-Consuming
  Tasks?}
\newblock \emph{arXiv preprint arXiv:2407.15711}, 2024.
\newblock URL \url{https://arxiv.org/abs/2407.15711}.

\bibitem[Drouin et~al.(2024)Drouin, Gasse, Caccia, Laradji, Del~Verme, Marty,
  Boisvert, Thakkar, Cappart, Vazquez, Chapados, and
  Lacoste]{drouin2024workarena}
Alexandre Drouin, Maxime Gasse, Massimo Caccia, Issam~H. Laradji, Manuel
  Del~Verme, Tom Marty, L{\'{e}}o Boisvert, Megh Thakkar, Quentin Cappart,
  David Vazquez, Nicolas Chapados, and Alexandre Lacoste.
\newblock {WorkArena: How Capable Are Web Agents at Solving Common Knowledge
  Work Tasks?}
\newblock \emph{arXiv preprint arXiv:2403.07718}, 2024.
\newblock URL \url{https://arxiv.org/abs/2403.07718}.

\bibitem[Xu et~al.(2024)Xu, Song, Li, Tang, Jain, Bao, Wang, Zhou, Guo, Cao,
  Yang, Lu, Martin, Su, Maben, Mehta, Chi, Jang, Xie, Zhou, and
  Neubig]{xu2024theagentcompany}
Frank~F. Xu, Yufan Song, Boxuan Li, Yuxuan Tang, Kritanjali Jain, Mengxue Bao,
  Zora~Z. Wang, Xuhui Zhou, Zhitong Guo, Murong Cao, Mingyang Yang, Hao~Yang
  Lu, Amaad Martin, Zhe Su, Leander Maben, Raj Mehta, Wayne Chi, Lawrence Jang,
  Yiqing Xie, Shuyan Zhou, and Graham Neubig.
\newblock {TheAgentCompany: Benchmarking LLM Agents on Consequential Real World
  Tasks}.
\newblock \emph{arXiv preprint arXiv:2412.14161}, 2024.
\newblock URL \url{https://arxiv.org/abs/2412.14161}.

\bibitem[Chan et~al.(2024)Chan, Chowdhury, Jaffe, Aung, Sherburn, Mays,
  Starace, Liu, Maksin, Patwardhan, Weng, and M{\k{a}}dry]{chan2024mlebench}
Jun~Shern Chan, Neil Chowdhury, Oliver Jaffe, James Aung, Dane Sherburn, Evan
  Mays, Giulio Starace, Kevin Liu, Leon Maksin, Tejal Patwardhan, Lilian Weng,
  and Aleksander M{\k{a}}dry.
\newblock {MLE-bench: Evaluating Machine Learning Agents on Machine Learning
  Engineering}.
\newblock \emph{arXiv preprint arXiv:2410.07095}, 2024.
\newblock URL \url{https://arxiv.org/abs/2410.07095}.

\bibitem[Chen et~al.(2024)Chen, Chen, Ning, Zhang, Wang, Yu, Li, Liao, Wei, Lu,
  Dey, Xue, Baker, Burns, Adu-Ampratwum, Huang, Ning, Gao, Su, and
  Sun]{chen2024scienceagentbench}
Ziru Chen, Shijie Chen, Yuting Ning, Qianheng Zhang, Boshi Wang, Botao Yu,
  Yifei Li, Zeyi Liao, Chen Wei, Zitong Lu, Vishal Dey, Mingyi Xue, Frazier~N.
  Baker, Benjamin Burns, Daniel Adu-Ampratwum, Xuhui Huang, Xia Ning, Song Gao,
  Yu~Su, and Huan Sun.
\newblock {ScienceAgentBench: Toward Rigorous Assessment of Language Agents for
  Data-Driven Scientific Discovery}.
\newblock \emph{arXiv preprint arXiv:2410.05080}, 2024.
\newblock URL \url{https://arxiv.org/abs/2410.05080}.

\bibitem[Yao et~al.(2024)Yao, Shinn, Razavi, and Narasimhan]{yao2024taubench}
Shunyu Yao, Noah Shinn, Pedram Razavi, and Karthik Narasimhan.
\newblock {$\tau$-bench: A Benchmark for Tool-Agent-User Interaction in
  Real-World Domains}.
\newblock \emph{arXiv preprint arXiv:2406.12045}, 2024.
\newblock URL \url{https://arxiv.org/abs/2406.12045}.

\bibitem[Barres et~al.(2025)Barres, Dong, Ray, Si, and
  Narasimhan]{barres2025tau2bench}
Victor Barres, Honghua Dong, Soham Ray, Xujie Si, and Karthik Narasimhan.
\newblock {$\tau^2$-Bench: Evaluating Conversational Agents in a Dual-Control
  Environment}.
\newblock \emph{arXiv preprint arXiv:2506.07982}, 2025.
\newblock URL \url{https://arxiv.org/abs/2506.07982}.

\bibitem[Hong et~al.(2023)Hong, Zhuge, Chen, Zheng, Cheng, Zhang, Wang, Wang,
  Yau, Lin, Zhou, Ran, Xiao, Wu, and Schmidhuber]{hong2023metagpt}
Sirui Hong, Mingchen Zhuge, Jiaqi Chen, Xiawu Zheng, Yuheng Cheng, Ceyao Zhang,
  Jinlin Wang, Zili Wang, Steven Ka~Shing Yau, Zijuan Lin, Liyang Zhou, Chenyu
  Ran, Lingfeng Xiao, Chenglin Wu, and J{\"{u}}rgen Schmidhuber.
\newblock {MetaGPT: Meta Programming for A Multi-Agent Collaborative
  Framework}.
\newblock \emph{arXiv preprint arXiv:2308.00352}, 2023.
\newblock URL \url{https://arxiv.org/abs/2308.00352}.

\bibitem[Qian et~al.(2023)Qian, Liu, Liu, Chen, Dang, Li, Yang, Chen, Su, Cong,
  Xu, Li, Liu, and Sun]{qian2023chatdev}
Chen Qian, Wei Liu, Hongzhang Liu, Nuo Chen, Yufan Dang, Jiahao Li, Cheng Yang,
  Weize Chen, Yusheng Su, Xin Cong, Juyuan Xu, Dahai Li, Zhiyuan Liu, and
  Maosong Sun.
\newblock {ChatDev: Communicative Agents for Software Development}.
\newblock \emph{arXiv preprint arXiv:2307.07924}, 2023.
\newblock URL \url{https://arxiv.org/abs/2307.07924}.

\bibitem[Wu et~al.(2023)Wu, Bansal, Zhang, Wu, Li, Zhu, Jiang, Zhang, Zhang,
  Liu, Awadallah, White, Burger, and Wang]{wu2023autogen}
Qingyun Wu, Gagan Bansal, Jieyu Zhang, Yiran Wu, Beibin Li, Erkang Zhu,
  Li~Jiang, Xiaoyun Zhang, Shaokun Zhang, Jiale Liu, Ahmed~Hassan Awadallah,
  Ryen~W White, Doug Burger, and Chi Wang.
\newblock {AutoGen: Enabling Next-Gen LLM Applications via Multi-Agent
  Conversation}.
\newblock \emph{arXiv preprint arXiv:2308.08155}, 2023.
\newblock URL \url{https://arxiv.org/abs/2308.08155}.

\bibitem[Cemri et~al.(2025)Cemri, Pan, Yang, Agrawal, Chopra, Tiwari, Keutzer,
  Parameswaran, Klein, Ramchandran, Zaharia, Gonzalez, and
  Stoica]{cemri2025mast}
Mert Cemri, Melissa~Z. Pan, Shuyi Yang, Lakshya~A. Agrawal, Bhavya Chopra,
  Rishabh Tiwari, Kurt Keutzer, Aditya Parameswaran, Dan Klein, Kannan
  Ramchandran, Matei Zaharia, Joseph~E. Gonzalez, and Ion Stoica.
\newblock {Why Do Multi-Agent LLM Systems Fail?}
\newblock \emph{arXiv preprint arXiv:2503.13657}, 2025.
\newblock URL \url{https://arxiv.org/abs/2503.13657}.

\bibitem[Zhuge et~al.(2024)Zhuge, Wang, Kirsch, Faccio, Khizbullin, and
  Schmidhuber]{zhuge2024gptswarm}
Mingchen Zhuge, Wenyi Wang, Louis Kirsch, Francesco Faccio, Dmitrii Khizbullin,
  and J{\"{u}}rgen Schmidhuber.
\newblock {Language Agents as Optimizable Graphs}.
\newblock \emph{arXiv preprint arXiv:2402.16823}, 2024.
\newblock URL \url{https://arxiv.org/abs/2402.16823}.

\bibitem[Hu et~al.(2024)Hu, Lu, and Clune]{hu2024adas}
Shengran Hu, Cong Lu, and Jeff Clune.
\newblock {Automated Design of Agentic Systems}.
\newblock \emph{arXiv preprint arXiv:2408.08435}, 2024.
\newblock URL \url{https://arxiv.org/abs/2408.08435}.

\bibitem[Zhang et~al.(2024)Zhang, Xiang, Yu, Teng, Chen, Chen, Zhuge, Cheng,
  Hong, Wang, Zheng, Liu, Luo, and Wu]{zhang2024aflow}
Jiayi Zhang, Jinyu Xiang, Zhaoyang Yu, Fengwei Teng, Xionghui Chen, Jiaqi Chen,
  Mingchen Zhuge, Xin Cheng, Sirui Hong, Jinlin Wang, Bingnan Zheng, Bang Liu,
  Yuyu Luo, and Chenglin Wu.
\newblock {AFlow: Automating Agentic Workflow Generation}.
\newblock \emph{arXiv preprint arXiv:2410.10762}, 2024.
\newblock URL \url{https://arxiv.org/abs/2410.10762}.

\bibitem[Shang et~al.(2024)Shang, Li, Zhao, Ma, Liu, Xu, and
  Li]{shang2024agentsquare}
Yu~Shang, Yu~Li, Keyu Zhao, Likai Ma, Jiahe Liu, Fengli Xu, and Yong Li.
\newblock {AgentSquare: Automatic LLM Agent Search in Modular Design Space}.
\newblock \emph{arXiv preprint arXiv:2410.06153}, 2024.
\newblock URL \url{https://arxiv.org/abs/2410.06153}.

\bibitem[Zhang et~al.(2025)Zhang, Niu, Fang, Wang, Bai, and
  Wang]{zhang2025maas}
Guibin Zhang, Luyang Niu, Junfeng Fang, Kun Wang, Lei Bai, and Xiang Wang.
\newblock {Multi-agent Architecture Search via Agentic Supernet}.
\newblock \emph{arXiv preprint arXiv:2502.04180}, 2025.
\newblock URL \url{https://arxiv.org/abs/2502.04180}.

\bibitem[Lin et~al.(2026)Lin, Yang, and Qin]{lin2026masloop}
Hehai Lin, Qi~Yang, and Chengwei Qin.
\newblock {MAS-Loop: Beyond One-Shot Orchestration towards Self-Evolving
  Multi-Agent Systems}.
\newblock \emph{arXiv preprint arXiv:2610.38946}, 2026.
\newblock URL \url{https://arxiv.org/abs/2610.38946}.

\bibitem[Liu et~al.(2023{\natexlab{b}})Liu, Zhang, Li, Liu, and
  Yang]{liu2023dylan}
Zijun Liu, Yanzhe Zhang, Peng Li, Yang Liu, and Diyi Yang.
\newblock {A Dynamic LLM-Powered Agent Network for Task-Oriented Agent
  Collaboration}.
\newblock \emph{arXiv preprint arXiv:2310.02170}, 2023{\natexlab{b}}.
\newblock URL \url{https://arxiv.org/abs/2310.02170}.

\bibitem[Xu et~al.(2023)Xu, Peng, Lei, Mukherjee, Liu, and Xu]{xu2023rewoo}
Binfeng Xu, Zhiyuan Peng, Bowen Lei, Subhabrata Mukherjee, Yuchen Liu, and
  Dongkuan Xu.
\newblock {ReWOO: Decoupling Reasoning from Observations for Efficient
  Augmented Language Models}.
\newblock \emph{arXiv preprint arXiv:2305.18323}, 2023.
\newblock URL \url{https://arxiv.org/abs/2305.18323}.

\bibitem[Li et~al.(2026)Li, Liu, Chen, You, Di, He, Zheng, Choe, Sun, Wang,
  Tao, Li, Zhao, Geng, Wu, Zhou, Chen, Xing, Li, Zeng, Wang, Wang, Chaim,
  Jiang, Shen, Kong, Liu, Wang, Liu, Li, Lan, Lin, Ye, He, Li, Zhang, Gao, Li,
  Ma, Jing, Wang, Li, Xue, Lyu, He, Tian, Wu, Wang, Gao, Chen, Liu, Cheng, Bao,
  Tong, Xu, Zhuo, Ye, Qi, Li, Liao, Tan, Shi, Tang, Tankasala, Yuan, Qian, Tu,
  Wang, Sun, Wang, Taylor, Yang, Guan, Dong, Zhang, Dillmann, chung Lee, and
  Song]{li2026skillsbench}
Xiangyi Li, Yimin Liu, Wenbo Chen, Bingran You, Zonglin Di, Yifeng He, Shenghan
  Zheng, Kyoung~Whan Choe, Jiankai Sun, Shuyi Wang, Chujun Tao, Binxu Li,
  Xuandong Zhao, Hejia Geng, Xiaojun Wu, Junwei Zhou, Xiaokun Chen, Hanwen
  Xing, Yubo Li, Qunhong Zeng, Di~Wang, Yuanli Wang, Roey~Ben Chaim, Penghao
  Jiang, Haotian Shen, Luyang Kong, Xinyi Liu, Runhui Wang, Xuanqing Liu,
  Jiachen Li, Xin Lan, Yueqian Lin, Wengao Ye, Junwei He, Songlin Li, Yue
  Zhang, Yipeng Gao, Yijiang Li, Ze~Ma, Liqiang Jing, Tianyu Wang, Kaixin Li,
  Yiqi Xue, Haoran Lyu, Yizhuo He, Yuchen Tian, Shutong Wu, Bowei Wang, Yixuan
  Gao, Bo~Chen, Litong Liu, Sikai Cheng, Jiajun Bao, Shuaicheng Tong, Shuwen
  Xu, Terry~Yue Zhuo, Tinghan Ye, Qi~Qi, Miao Li, Longtai Liao, Zelin Tan,
  Chang Shi, Xilin Tang, Srinath Tankasala, Boqin Yuan, Yaoyao Qian, Jianhong
  Tu, Chenguang Wang, Yizhou Sun, Wei Wang, Aaron Taylor, Ziyue Yang, Changkun
  Guan, Zhikang Dong, Xinyu Zhang, Steven Dillmann, Han chung Lee, and Dawn
  Song.
\newblock {SkillsBench: Benchmarking How Well Agent Skills Work Across Diverse
  Tasks}.
\newblock \emph{arXiv preprint arXiv:2602.12670}, 2026.
\newblock URL \url{https://arxiv.org/abs/2602.12670}.

\bibitem[Xia et~al.(2024)Xia, Deng, Dunn, and Zhang]{xia2024agentless}
Chunqiu~Steven Xia, Yinlin Deng, Soren Dunn, and Lingming Zhang.
\newblock {Agentless: Demystifying LLM-based Software Engineering Agents}.
\newblock \emph{arXiv preprint arXiv:2407.01489}, 2024.
\newblock URL \url{https://arxiv.org/abs/2407.01489}.

\end{thebibliography}
\endgroup

\clearpage
\appendix

\section{Detailed Task Catalog}
\label{app:task_catalog}

Table~\ref{tab:suite} describes each task form: its source material, how the task is instantiated, and the evidence by which the final deliverable is accepted.

{\footnotesize
\setlength{\tabcolsep}{4pt}
\begin{longtable}{@{}>{\raggedright\arraybackslash}p{.15\linewidth}r>{\raggedright\arraybackslash}p{.21\linewidth}>{\raggedright\arraybackslash}p{.31\linewidth}>{\raggedright\arraybackslash}p{.19\linewidth}@{}}
\caption{WorkflowArena task forms: \BenchmarkTasks{} tasks across \BenchmarkDomains{} domains and \BenchmarkForms{} task forms. For each form, the table lists its source material, how the task is instantiated, and the evidence by which the final deliverable is accepted.}\label{tab:suite}\\
\toprule Task form & Tasks & Source material & Instantiation and integration requirement & Acceptance evidence \\ \midrule
\endfirsthead
\caption[]{(continued)}\\
\toprule Task form & Tasks & Source material & Instantiation and integration requirement & Acceptance evidence \\ \midrule
\endhead
\multicolumn{5}{r}{\emph{Continued on next page}}\\
\endfoot
\bottomrule\endlastfoot
\multicolumn{5}{@{}l}{\textbf{SWE} (17 tasks)} \\*[1pt]
Feature implementation & 11 & Open-source repository revisions and pull requests. & Implement a specified feature across an existing codebase while preserving unrelated behavior. & Hidden feature and regression tests \\[1pt]
Subsystem reimplementation & 6 & Versioned open-source subsystems. & Remove implementation bodies while retaining the public interface and surrounding repository; reconstruct the missing behavior. & Behavior recovery and regression tests \\[1pt]
\midrule
\multicolumn{5}{@{}l}{\textbf{OPS} (18 tasks)} \\*[1pt]
Policy-constrained process execution & 6 & Public expenditure records with authored policies and exceptions. & Execute a queue of approvals or transactions under authorization, ordering, and resource constraints. & Outcome, action history, constraints \\[1pt]
Cross-system business orchestration & 6 & Authored business scenarios grounded in operational records. & Coordinate records and obligations across business systems, including identity, access, finance, and case handling. & Final states and cross-system consistency \\[1pt]
Multi-party information coordination & 6 & Authored stakeholder roles, information access, and handover requirements. & Acquire distributed evidence and reconcile it into a final handover with traceable sources. & Evidence acquisition, process, handover \\[1pt]
\midrule
\multicolumn{5}{@{}l}{\textbf{SCI} (16 tasks)} \\*[1pt]
Blind predictive modeling & 8 & CAPRI, CASP, DREAM, SAMPL, and molecular-property challenge data. & Fit or design a predictor from public inputs and submit predictions for hidden evaluation targets. & Official metric and reference ranking \\[1pt]
Published result reproduction & 8 & Published scientific studies and their data or computational inputs. & Reconstruct prescribed quantitative results with the stated conventions and execution requirements. & Numerical targets within tolerances \\[1pt]
\midrule
\multicolumn{5}{@{}l}{\textbf{MATH} (24 tasks)} \\*[1pt]
Formal proof construction & 16 & Lean formalizations of mathematical results. & Complete target proofs in a dependency-bearing development without changing the theorem statements. & Lean type checking and axiom audit \\[1pt]
Formal API development & 8 & Families of mathlib4 developments. & Implement a related collection of declarations and integrate them with the surrounding formal library. & Required declarations and library build \\[1pt]
\midrule
\multicolumn{5}{@{}l}{\textbf{FIN} (16 tasks)} \\*[1pt]
Cross-filing peer analysis & 10 & Frozen company filing corpora and authored analytical contracts. & Reconcile periods, entities, and accounting conventions before deriving comparisons and sensitivity certificates. & Numerical, evidence, and synthesis checks \\[1pt]
Multi-entity consolidation and audit & 6 & Simulated multi-entity ledgers and authored accounting events. & Reconstruct adjustments and consolidate or audit entities with currency, acquisition, and intercompany dependencies. & Ledger, adjustment, and reconciliation checks \\[1pt]
\midrule
\multicolumn{5}{@{}l}{\textbf{LAW} (22 tasks)} \\*[1pt]
Contract portfolio analysis & 13 & Merger-agreement portfolios. & Analyze individual clauses and synthesize cross-contract relations, exceptions, and conflicts. & Clause-level and portfolio-level checks \\[1pt]
Multi-party case adjudication & 9 & Chinese criminal-judgment materials and authored case requirements. & Resolve participant-level findings and produce consistent joint legal and sentencing conclusions. & Individual and joint disposition checks \\[1pt]
\midrule
\multicolumn{5}{@{}l}{\textbf{CDN} (14 tasks)} \\*[1pt]
Narrative video assembly & 8 & Source films, editing projects, scripts, and accessibility materials. & Reconcile edits, narrative timing, audio, captions, and requested project deliverables into a coherent release. & Media, editorial, and project rubric \\[1pt]
Presentation and document design & 6 & Public technical, incident, and scientific reports. & Transform long source material into an audience-specific presentation or document with grounded claims. & Content, evidence, and delivery rubric \\[1pt]
\midrule
\multicolumn{5}{@{}l}{\textbf{3D} (18 tasks)} \\*[1pt]
3D scene construction & 10 & Construction drawings, site information, and authored change orders. & Build a Blender scene whose components satisfy shared spatial interfaces and render requirements. & Geometry, placement, and delivery rubric \\[1pt]
3D narrative short film & 8 & Story briefs, scripts, and scene assets. & Produce coordinated shots, actions, and transitions as a complete rendered film. & Shots, timing, continuity, and delivery rubric \\[1pt]
\midrule
Total & \BenchmarkTasks{} & \multicolumn{3}{l}{\BenchmarkForms{} task forms across \BenchmarkDomains{} domains} \\
\end{longtable}
}

\section{Task Package Format}
\label{app:package}

Task packages follow the Harbor execution interface. Each package contains an agent-facing instruction (\texttt{instruction.md}), an environment specification (\texttt{environment/}), a verifier entry point (\texttt{tests/}), and task metadata. The instruction defines the required deliverable and execution constraints, while the environment provides the initial workspace and task inputs. Hidden tests and reference assets are kept separate from the agent-facing materials. After execution, the verifier evaluates the final submission and any task-relevant state transitions to produce the task score.

\section{Task-Specific Scoring Criteria}
\label{app:scoring}

Each task is evaluated with a continuous score on a 0--100 scale. The verifier assesses the submitted deliverable against the requirements in the task contract. Component checks contribute to this score through the task-specific scoring rule.

Software and formal mathematics scores reflect progress across functional groups or required targets, together with build and integrity checks. Operational scores combine outcome, process, and constraint checks. Financial and legal scores account for both local analyses and their integration into a consistent final report. Scientific reproduction uses target-specific numerical tolerances, while predictive modeling uses the official metric and a fixed reference population. Content and 3D scores use weighted rubrics covering substantive requirements and delivery quality, including narrative coherence, geometry, and cross-component consistency where applicable; these rubrics are scored by Gemini-3.8-Flash.

The task contract specifies the scoring weights, tolerances, equivalence rules, and required checks. Artifact evaluation combines technical inspection with semantic review: media must satisfy the requested narrative and editing requirements, and scenes must satisfy the specified geometry and spatial relationships. The resulting score captures the extent to which the final deliverable fulfills the task.

\section{Execution and Resource Accounting}
\label{app:execution}

\subsection{Model and execution configuration}

Table~\ref{tab:execution-prices} lists the API model identifiers and harness versions. The main agent and its workers use the same registered model, harness, and effort. Model identifiers are passed to the API through the execution broker, which also enforces the effort setting. Codex is pinned to version 0.164.0 and Claude Code to version 2.1.280. The trial registration records the agent profile, protocol options, task-package hash, and container image identifiers. At execution start, the runner archives its Python implementation, prompt templates, JavaScript runtime, and interface schemas with SHA-256 hashes. These artifacts identify the executable configuration of each trial independently of subsequent repository changes.

\paragraph{Containers and resources.}
Task execution takes place in Docker containers, with model inference accessed through the broker. The task package specifies CPU and memory limits for each agent container; these override the generic agent-profile defaults. The runner records the effective limits, CPU affinity when applicable, and resolved image identifier in the execution record. A host-level admission controller reserves CPU and memory before starting a container and holds the reservation until cleanup. Its capacity is bounded by 90\% of the host's reported CPUs and 85\% of its memory, with any explicitly configured host caps applied as additional limits. The resulting capacity and container reservations are recorded alongside the trial. Thus, task-level resource declarations and effective container allocations specify the compute available for shell commands, tests, rendering, and other local work.

\paragraph{Wall-clock accounting and request handling.}
The four-hour protocol uses a 14,400-second deadline shared by the main agent and all workers. The execution loop establishes this deadline using a monotonic clock and passes the remaining time to each agent invocation. Worker dispatch, container admission waits, tool execution, provider queueing, and retry backoff consume the same budget; launching or resuming a Workflow does not reset it. Task preparation precedes this execution window, and grading follows it. Worker concurrency is capped at 16. The broker separately limits simultaneous upstream requests to the registered concurrency and applies a 240-second queue timeout.

Transient transport failures, rate-limit responses, and provider overloads are handled by the broker. Backoff grows exponentially, is capped at 30 seconds, and adds up to 0.3 seconds of jitter; a valid provider \texttt{Retry-After} value of at most 120 seconds can extend the wait. Retries remain bounded by the request deadline and the transport policy in the archived implementation. The broker records the attempt limit, response status, queue time, elapsed time, and backoff for each request. Native harness request retries are disabled when broker-managed retries are enabled, avoiding nested retry loops. Delivered model or tool events are not replayed after a stream failure. Provider-side service capacity is external to the container resource allocation; its observed waits are included in the execution time.

\subsection{Prompts and tool interfaces}

The initial message to the main agent consists of three parts. The first sets up the task. It points the agent to the task instruction in \path{/testbed/instruction.md}, states that the workspace is writable, and allows the agent to work directly, to delegate through Workflows, or to combine the two, without a required planning stage and without prescribed roles, topology, or worker count. It asks the agent to prioritize correctness and completeness before elapsed time and token consumption, to coordinate file ownership among workers that share the workspace, to give each worker the context it needs because workers start from fresh conversations, and to confirm that the final deliverable is in place before finishing. It also reports the total and remaining time budget and notes that launching a Workflow does not reset it. The second part defines Workflow programs, including their script format and the interface described below, and advises that the structure of a workflow should follow the task rather than a fixed template of stages. The third part describes runtime details, such as where execution transcripts can be inspected and how changes made in isolated worktrees must be integrated. Each subsequent turn begins with a short continuation message that reports execution feedback and the remaining budget.

The harness retains its versioned native instructions. Benchmark messages and worker assignments are supplied explicitly; automatic discovery of host project instructions, memories, and skills is disabled. Workflow workers start with fresh conversations and receive the assignment and public task workspace. Optional role instructions use the harness instruction channel: Codex's developer instructions or Claude Code's appended system prompt. The runner saves the rendered worker prompt and role-instruction hash. Workers share \path{/testbed} by default; isolated Git worktrees require explicit integration by the agent.

The main agent receives five Workflow interfaces: \texttt{Workflow} launches a background JavaScript program; \texttt{WorkflowList} lists runs and active workers; \texttt{WorkflowOutput} retrieves status or results and can wait for up to 30 seconds; \texttt{WorkflowStop} cancels a run and its workers; and \texttt{WorkflowControl} pauses or resumes dispatch, or stops or restarts an individual worker. Their JSON schemas are supplied through MCP. Workflow programs compose \texttt{agent}, \texttt{parallel}, and \texttt{pipeline} calls, the primitives of Claude Code dynamic workflows~\citep{anthropic2026dynamicworkflows}, with ordinary JavaScript control flow; both harnesses receive the same interface. Environment operations are performed through worker tools. Claude Code exposes reading, searching, shell execution, and file editing; Codex exposes the corresponding native workspace tools and local image inspection. Native sub-agent tools are disabled, so delegation passes through the recorded Workflow interface.

\subsection{Token categories and inference-cost calculation}

For each task, usage is summed over the main agent and Workflow workers, including recorded retry attempts. We normalize usage into four disjoint categories: uncached input $I$, cache reads $R$, cache writes $W$, and output $O$. For Responses-style records, the reported input count already includes cached input, so $I=T_{\mathrm{input}}-R-W$. For Messages-style records, the reported input count excludes cache reads and writes, so $I=T_{\mathrm{input}}$. In both cases, total token consumption is $I+R+W+O$. Reasoning tokens already included in output are not added again, and audit summaries of the same calls are not counted as additional usage.

For model $m$, the estimated inference cost of task $t$ is
\begin{equation}
    C_{m,t}=10^{-6}\bigl(p_{m,I}I_{m,t}+p_{m,R}R_{m,t}
    +p_{m,W}W_{m,t}+p_{m,O}O_{m,t}\bigr),
\end{equation}
where prices are in US dollars per million tokens. Table~\ref{tab:execution-prices} gives the price snapshot used in the calculation. The estimate uses standard short-context pricing, five-minute cache-write pricing for Claude and Kimi, and the peak tariff for DeepSeek. For GLM, Doubao, and DeepSeek, cache creation is priced as uncached input. Doubao's CNY prices are converted at 6.7168 CNY per USD, using the exchange-rate snapshot of November 6, 2026. The estimate covers token-based model inference; grading, tools, container infrastructure, taxes, and time-based cache storage are outside this quantity.

\begin{table}[t]
\centering
\caption{API identifiers, harnesses, and inference-price snapshot dated November 9, 2026. Prices are USD per million tokens. Doubao prices are converted from CNY at 6.7168 CNY per USD.}
\label{tab:execution-prices}
\small
\setlength{\tabcolsep}{5pt}
\begin{tabular}{llrrrr}
\toprule
API model identifier & Harness & Uncached input & Cache read & Cache write & Output \\
\midrule
\texttt{claude-opus-5-5}       & Claude Code & 4.00 & 0.20 & 5.00 & 20.00 \\
\texttt{claude-sonnet-5-5}     & Claude Code & 2.00 & 0.10 & 2.50 & 10.00 \\
\texttt{gpt-6-astra}           & Codex & 10.00 & 1.00 & 12.50 & 50.00 \\
\texttt{gpt-6.1-sol}           & Codex & 2.00 & 0.10 & 2.50 & 10.00 \\
\texttt{gpt-6-luna}            & Codex & 0.10 & 0.01 & 0.125 & 0.50 \\
\texttt{kimi-k3.1-pro}         & Claude Code & 2.00 & 0.20 & 2.00 & 10.00 \\
\texttt{glm-5.3-flash}         & Claude Code & 0.15 & 0.03 & 0.15 & 0.50 \\
\texttt{deepseek-v4.1-flash}   & Claude Code & 0.30 & 0.006 & 0.30 & 1.20 \\
\texttt{doubao-seed-evolving}  & Claude Code & 0.893283 & 0.178657 & 0.893283 & 4.466413 \\
\bottomrule
\end{tabular}
\end{table}

\section{Workflow Transfer Setup}
\label{app:transfer}

The transfer subset contains 32 of the \BenchmarkTasks{} tasks and covers all \BenchmarkDomains{} domains; within each domain, we select the more difficult tasks. All settings in Table~\ref{tab:workflow-transfer} use this subset. In the Self setting, the execution model serves as its own workflow source under the rules of the dynamic mode, so the source model never edits the workspace in any setting.

\clearpage
\section{Acceptance Criteria and Resolved Rates}
\label{app:pass-rates}

Table~\ref{tab:pass-rate-criteria} lists the acceptance criteria that determine whether a task is resolved, as defined in \S\ref{sec:protocol}. Thresholds apply to each task, and critical checks are required in addition to any score threshold.

\begin{table}[H]
\centering
\caption{Acceptance criteria used to compute resolved rates.}
\label{tab:pass-rate-criteria}
\begingroup\small
\setlength{\tabcolsep}{5pt}
\renewcommand{\arraystretch}{1.12}
\begin{tabularx}{\textwidth}{@{}lX@{}}
\toprule
Domain & Acceptance criteria \\
\midrule
SWE & \textbf{Feature implementation:} successful build and import; all required tests pass; verified expected test coverage, allowing only task-defined legitimate skips; no test-integrity violations. \textbf{Subsystem reimplementation:} all designated fail-to-pass tests pass, all previously passing behaviors remain intact, and no integrity violations or unresolved grader errors. \\
\addlinespace[4pt]
OPS & Business-goal check accuracy $\geq90\%$, process check accuracy $\geq90\%$, and all constraint checks pass. Each dimension uses its own check count as the denominator. \\
\addlinespace[4pt]
SCI & \textbf{Blind predictive modeling:} valid submissions with required target coverage, meeting the task-specific reference baseline under its prescribed metric comparison. \textbf{Published result reproduction:} all numerical targets meet their tolerances, with verified code and execution evidence supporting the reproduction. \\
\addlinespace[4pt]
MATH & Successful Lean build; all required declarations have the prescribed types; permitted-axiom and dependency restrictions hold; task-pinned definition values match. Applies to both formal proof construction and formal API development. \\
\addlinespace[4pt]
FIN & \textbf{Cross-filing peer analysis:} all required root checks pass. \textbf{Multi-entity consolidation and audit:} native full-success conditions hold, including required branch and aggregate comparisons and hard consistency checks. \\
\addlinespace[4pt]
LAW & \textbf{Contract portfolio analysis:} clause-extraction score $\geq90\%$ and aggregation-question exact correctness $\geq75\%$. \textbf{Multi-party case adjudication:} correct legal determinations and satisfaction of the required sentencing and consistency criteria. \\
\addlinespace[4pt]
CDN & \textbf{Presentation and document design:} score $\geq75$. \textbf{Narrative video assembly:} score $\geq90$. Both require all critical rubric items to pass. \\
\addlinespace[4pt]
3D & \textbf{3D scene construction and 3D narrative short film:} score $\geq80$ and all critical rubric items pass. \\
\bottomrule
\end{tabularx}
\endgroup
\end{table}

\paragraph{Resolved rates versus continuous scores.}
The two overall measures order the 25 configurations similarly, with a Spearman rank correlation of 0.90 between Overall Resolved and Overall Score, but they diverge sharply within domains. Averaged over the 25 configurations, the continuous domain score exceeds the domain resolved rate by 11.23 points in MATH, 32.71 in CDN, 34.73 in SCI, 38.33 in OPS, 47.25 in 3D, 61.18 in FIN, 66.16 in SWE, and 67.76 in LAW. MATH is the only domain in which the two measures nearly coincide, because its acceptance criteria are close to the scoring rule itself: every required declaration must type-check under the prescribed axioms. In SWE, FIN, and LAW, by contrast, submissions routinely earn most of the partial credit while missing at least one acceptance criterion. Opus at high effort scores 96.98 in SWE yet resolves 17.65\% of SWE tasks, and LAW averages a score of 70.7 across configurations but a resolved rate of only 2.9\%.

\paragraph{Partial progress and complete delivery can rank systems differently.}
The domain-level divergence also changes which systems appear strongest. In SWE, Astra and Sol score between 71 and 79 in every configuration but resolve no SWE task except Astra at low effort, whereas Opus configurations resolve up to 29.41\%. Sonnet at max effort ranks fifteenth by Overall Score but fourth by Overall Resolved, and it attains the highest resolved rate in CDN despite a CDN score below that of two Opus configurations. Conversely, Sol at max effort leads OPS by continuous score with 72.97 but resolves 22.22\% of OPS tasks, below the 27.78\% of Astra at high effort and Doubao. Reporting both measures therefore distinguishes systems that make broad partial progress from those that more often deliver a fully accepted artifact.

\section{Continuous Scores}
\label{app:score-resources}

\begin{figure}[p]
\centering
\includegraphics[width=.9\linewidth]{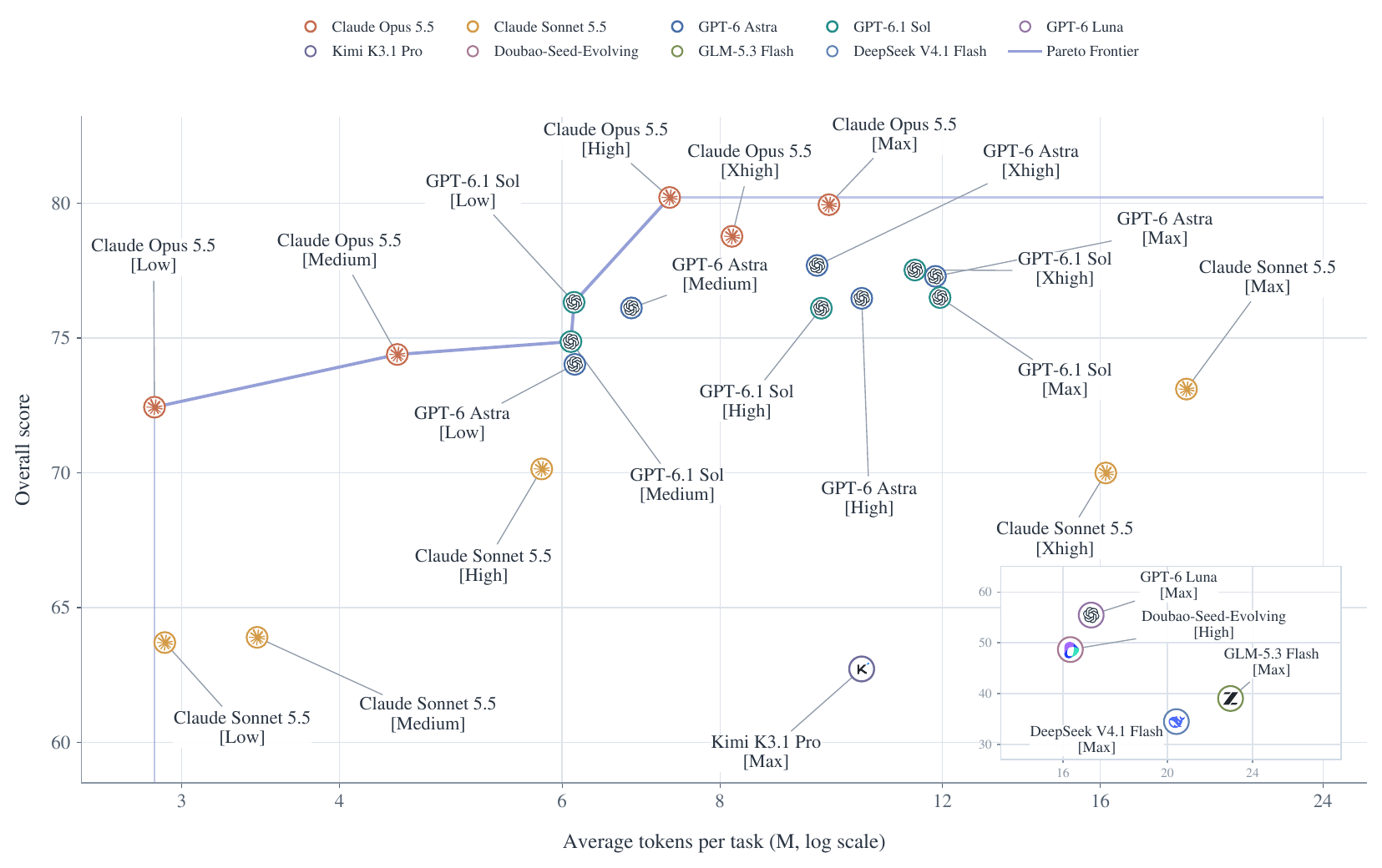}
\caption{\textbf{Overall Score versus token consumption.} Scores match Table~\ref{tab:scores}. Mean per-task tokens include main-agent and worker usage. The token axis is logarithmic, and the inset uses its own axes. The line connects non-dominated configurations.}
\label{fig:score-tokens}

\vspace{1.2em}
\includegraphics[width=.9\linewidth]{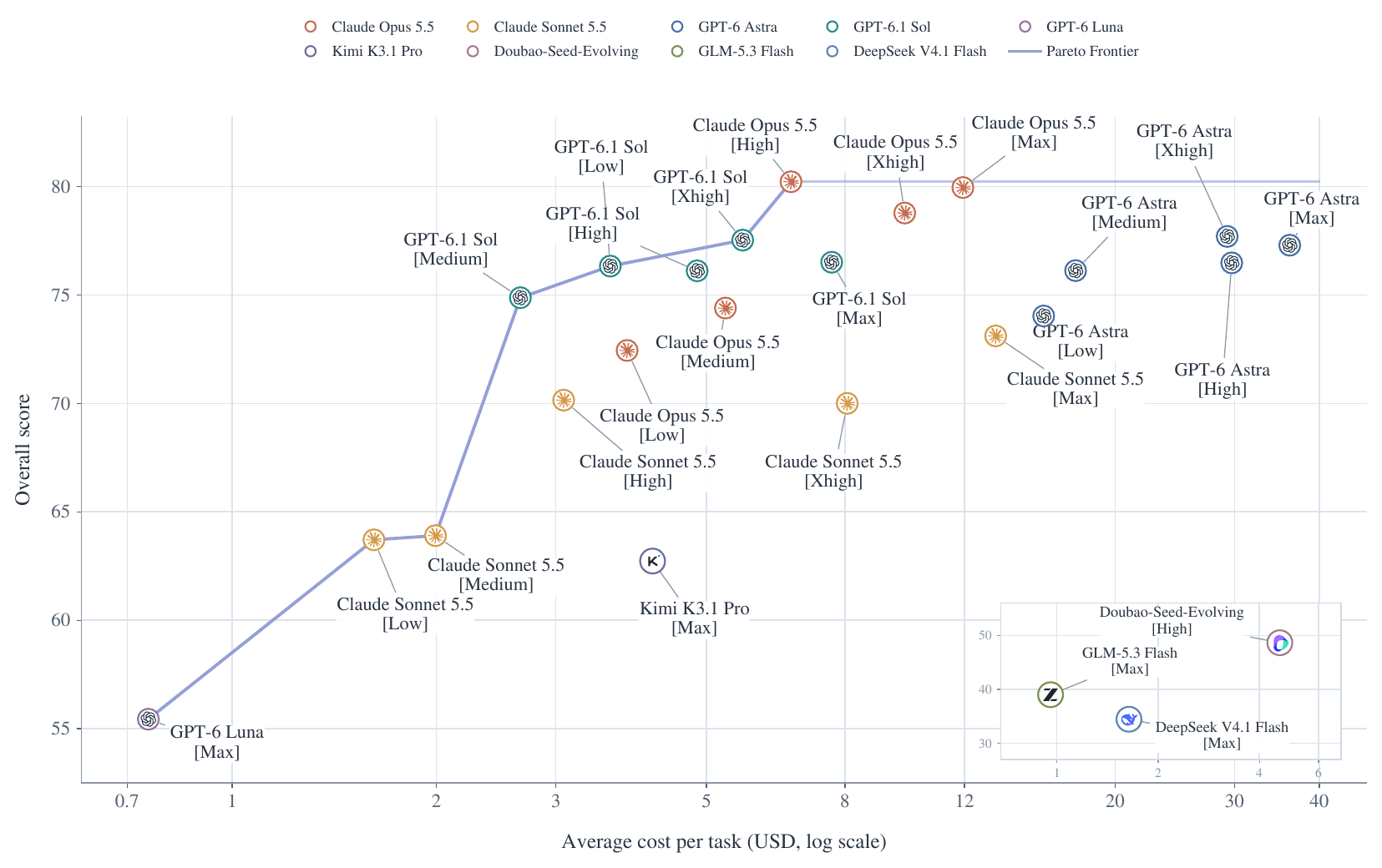}
\caption{\textbf{Overall Score versus mean inference cost per task.} Scores match Table~\ref{tab:scores}. Costs include the main agent and workers and are estimated from recorded token usage and model-specific API prices. The line connects non-dominated configurations.}
\label{fig:score-cost}
\end{figure}

Table~\ref{tab:scores} reports the continuous task scores defined in \S\ref{sec:protocol}, which credit partial progress toward each deliverable. Figures~\ref{fig:score-tokens} and~\ref{fig:score-cost} pair the Overall Score with the token and cost measurements of Figures~\ref{fig:resolved-tokens} and~\ref{fig:resolved-cost}.

\begin{table*}[htbp]
\centering\begingroup\footnotesize
\setlength{\tabcolsep}{2.5pt}\renewcommand{\arraystretch}{1.28}
\setlength{\aboverulesep}{0pt}\setlength{\belowrulesep}{0pt}
\colorlet{mainoverall}{brandprimary!13}
\newcommand{\modellogo}[1]{\raisebox{-0.18em}{\includegraphics[width=1.1em,height=1.1em,keepaspectratio]{Figures/model-logos/#1.pdf}}\hspace{0.35em}}
\caption{Continuous task scores across eight domains (0--100), which credit partial progress and complement the resolved rates in Table~\ref{tab:main}. Overall is the equal-weight mean of the domain scores, and rows are sorted by Overall. Bold and underlined values denote the highest and second-highest scores in each column.}
\label{tab:scores}
\resizebox{\textwidth}{!}{%
\begin{tabular}{@{}lcc|>{\columncolor{mainoverall}}c|*{8}{c}@{}}
\toprule
Model & Harness & Effort & \multicolumn{1}{c|}{Overall} & SWE & OPS & SCI & MATH & FIN & LAW & CDN & 3D \\
\midrule
\modellogo{claude-color}Claude Opus 5.5 & Claude Code & high & \textbf{80.21} & \textbf{96.98} & 54.94 & \underline{87.97} & \underline{97.44} & 81.69 & 74.36 & 79.72 & 68.60 \\
\modellogo{claude-color}Claude Opus 5.5 & Claude Code & max & \underline{79.94} & 79.26 & 52.91 & \textbf{94.71} & 95.83 & \textbf{87.79} & \underline{74.63} & \underline{82.60} & \textbf{71.82} \\
\modellogo{claude-color}Claude Opus 5.5 & Claude Code & xhigh & 78.77 & \underline{91.10} & 50.03 & 82.03 & \textbf{97.67} & 84.29 & \textbf{74.64} & \textbf{83.13} & 67.27 \\
\modellogo{openai}GPT-6 Astra & Codex & xhigh & 77.69 & 74.24 & 62.65 & 82.18 & 96.96 & 84.01 & 73.58 & 81.86 & 66.05 \\
\modellogo{openai}GPT-6.1 Sol & Codex & xhigh & 77.52 & 76.11 & 65.68 & 82.21 & 95.35 & \underline{85.02} & 71.10 & 81.65 & 63.05 \\
\modellogo{openai}GPT-6 Astra & Codex & max & 77.29 & 74.67 & 64.97 & 78.78 & 97.20 & 82.30 & 73.30 & 80.14 & 66.93 \\
\modellogo{openai}GPT-6.1 Sol & Codex & max & 76.50 & 71.19 & \textbf{72.97} & 79.76 & 95.36 & 77.43 & 71.36 & 77.66 & 66.28 \\
\modellogo{openai}GPT-6 Astra & Codex & high & 76.47 & 77.85 & 64.84 & 78.74 & 95.84 & 80.92 & 70.64 & 79.08 & 63.87 \\
\modellogo{openai}GPT-6.1 Sol & Codex & low & 76.32 & 78.43 & 62.06 & 82.43 & 91.02 & 82.05 & 70.91 & 80.84 & 62.85 \\
\modellogo{openai}GPT-6 Astra & Codex & medium & 76.12 & 76.09 & 64.02 & 77.46 & 92.63 & 83.00 & 71.01 & 80.04 & 64.67 \\
\modellogo{openai}GPT-6.1 Sol & Codex & high & 76.11 & 76.16 & \underline{66.58} & 79.47 & 94.71 & 81.33 & 70.79 & 76.25 & 63.56 \\
\modellogo{openai}GPT-6.1 Sol & Codex & medium & 74.87 & 76.82 & 63.98 & 78.57 & 88.06 & 75.57 & 72.72 & 80.38 & 62.84 \\
\modellogo{claude-color}Claude Opus 5.5 & Claude Code & medium & 74.39 & 84.86 & 53.04 & 86.26 & 77.63 & 80.94 & 72.26 & 77.50 & 62.63 \\
\modellogo{openai}GPT-6 Astra & Codex & low & 74.02 & 71.64 & 63.28 & 77.82 & 83.11 & 80.05 & 71.99 & 80.88 & 63.42 \\
\modellogo{claude-color}Claude Sonnet 5.5 & Claude Code & max & 73.11 & 73.35 & 52.45 & 84.31 & 90.39 & 70.23 & 61.84 & 80.97 & \underline{71.32} \\
\modellogo{claude-color}Claude Opus 5.5 & Claude Code & low & 72.44 & 84.50 & 45.70 & 79.14 & 72.44 & 77.59 & 73.87 & 78.61 & 67.62 \\
\modellogo{claude-color}Claude Sonnet 5.5 & Claude Code & high & 70.15 & 69.87 & 45.61 & 77.36 & 79.58 & 70.71 & 71.61 & 79.80 & 66.61 \\
\modellogo{claude-color}Claude Sonnet 5.5 & Claude Code & xhigh & 70.00 & 80.18 & 37.95 & 82.11 & 81.00 & 58.01 & 73.75 & 81.79 & 65.18 \\
\modellogo{claude-color}Claude Sonnet 5.5 & Claude Code & medium & 63.90 & 67.96 & 43.01 & 72.45 & 70.56 & 49.67 & 71.40 & 74.33 & 61.82 \\
\modellogo{claude-color}Claude Sonnet 5.5 & Claude Code & low & 63.71 & 60.37 & 45.57 & 68.78 & 70.76 & 62.20 & 70.16 & 73.71 & 58.11 \\
\modellogo{kimi-color}Kimi K3.1 Pro & Claude Code & max & 62.73 & 52.79 & 50.35 & 74.10 & 54.37 & 67.26 & 74.14 & 76.94 & 51.86 \\
\modellogo{openai}GPT-6 Luna & Codex & max & 55.44 & 51.66 & 34.36 & 66.87 & 36.55 & 68.50 & 66.24 & 76.20 & 43.12 \\
\modellogo{doubao-color}Doubao-Seed-Evolving & Claude Code & high & 48.66 & 35.87 & 52.73 & 63.68 & 4.07 & 55.12 & 70.29 & 66.47 & 41.04 \\
\modellogo{zai}GLM-5.3 Flash & Claude Code & max & 39.03 & 25.75 & 50.60 & 54.99 & 7.61 & 57.13 & 63.90 & 21.56 & 30.71 \\
\modellogo{deepseek-color}DeepSeek V4.1 Flash & Claude Code & max & 34.48 & 34.45 & 43.48 & 52.25 & 10.36 & 46.81 & 56.28 & 0.00 & 32.25 \\
\bottomrule
\end{tabular}}\endgroup
\end{table*}

Continuous scores are considerably higher than resolved rates. Opus at high effort attains the highest Overall Score, 80.21, with 96.98 in SWE and 97.44 in MATH, even though the highest SWE resolved rate is 29.41\%. Opus configurations lead seven of the eight domains, at three different effort settings, while Sol at max effort leads OPS with 72.97. OPS separates the two model families most clearly under this measure: every Astra and Sol configuration scores above 62, whereas no Opus or Sonnet configuration exceeds 55.

Additional tokens yield smaller gains in continuous score than in resolved rate. Sonnet at max effort uses 3.2$\times$ the tokens of Sonnet at high effort for a gain of 2.96 points, from 70.15 to 73.11, whereas its resolved rate rises by 7.95 points; Sol at low effort comes within 1.20 points of its xhigh configuration while using 54\% of the tokens. Additional tokens therefore contribute more to completing tasks fully than to partial progress. With respect to cost, Sol at low effort scores 76.32 at \$3.61 per task and Opus at high effort scores 80.21 at \$6.66, whereas Astra at xhigh effort scores 77.69 at \$29.26.

\end{document}